\documentclass[10pt,twocolumn,letterpaper]{article}

\usepackage{cvpr}
\usepackage{times}
\usepackage{epsfig}
\usepackage{graphicx}
\usepackage{amsmath}
\usepackage{amssymb}
\usepackage{multirow}

\usepackage{url}
\def\myurl#1{\setbox0\vbox{\hsize.5\maxdimen
		\url{#1}\par
		\global\setbox1\lastbox}\unhbox1 }

\usepackage{caption}
\usepackage{subcaption}

\usepackage[pagebackref=true,breaklinks=true,letterpaper=true,colorlinks,bookmarks=false]{hyperref}

\newcommand{\bx} {{\bf x }}
\newcommand{\bW} {{\bf W }}

\newcommand{\bh} {{\bf h }}

\newcommand{\bb} {{\bf b }}

\newcommand{\by} {{\bf y }}

\newcommand{\bk} {{\bf k }}

\newcommand{\bz} {{\bf z }}
\newcommand{\bo} {{\bf o }}

\newcommand{\bla} {{\boldsymbol \lambda }}

\usepackage[lined,boxed,commentsnumbered]{algorithm2e}
\newcommand{\trans}[1]{{#1}^{\ensuremath{\mathsf{T}}}} 

\cvprfinalcopy 

\def\cvprPaperID{473} 

\ifcvprfinal\fi
\begin{document}

\title{Pedestrian Detection aided by Deep Learning Semantic Tasks}

\author{Yonglong Tian$^1$, Ping Luo$^{1}$, Xiaogang Wang$^{2}$, Xiaoou Tang$^{1}$ \\
$^1$Department of Information Engineering, The Chinese University of Hong Kong \\
$^2$Department of Electronic Engineering, The Chinese University of Hong Kong \\
\texttt{\{ty014,pluo\thanks{For more technical details of this work, please send email to \url{pluo.lhi@gmail.com}},xtang\}@ie.cuhk.edu.hk,~\{xgwang\}@ee.cuhk.edu.hk}
}

\maketitle

\begin{abstract}
Deep learning methods have achieved great success in pedestrian detection, owing to its ability to learn features from raw pixels. However,
they mainly capture middle-level representations, such as pose of pedestrian, but confuse positive with hard negative samples (Fig.\ref{fig:intro} (a)), which have large ambiguity, \eg the shape and appearance of `tree trunk' or `wire pole' are similar to pedestrian in certain viewpoint.
This ambiguity can be distinguished by high-level representation.
To this end, this work jointly optimizes pedestrian detection with semantic tasks, including pedestrian attributes (\eg `carrying backpack') and scene attributes (\eg `road', `tree', and `horizontal').
%
%
Rather than expensively annotating scene attributes, we transfer attributes information from existing scene segmentation datasets to the pedestrian dataset,
by proposing a novel deep model to learn high-level features from multiple tasks and multiple data sources.
Since distinct tasks have distinct convergence rates and data from different datasets have different distributions,
a multi-task objective function is carefully designed to coordinate tasks
and reduce discrepancies among datasets.
The importance coefficients of tasks and network parameters in this objective function can be iteratively estimated.
%
Extensive evaluations show that the proposed approach outperforms the state-of-the-art on the challenging Caltech \cite{Dollar2012PAMI} and ETH \cite{ess2007depth} datasets, where
it reduces the miss rates of previous deep models by 17 and 5.5 percent, respectively.
\end{abstract}

\section{Introduction}\label{sec:intro}

\begin{figure}[t]
	\centering
	\includegraphics[width=0.95\linewidth]{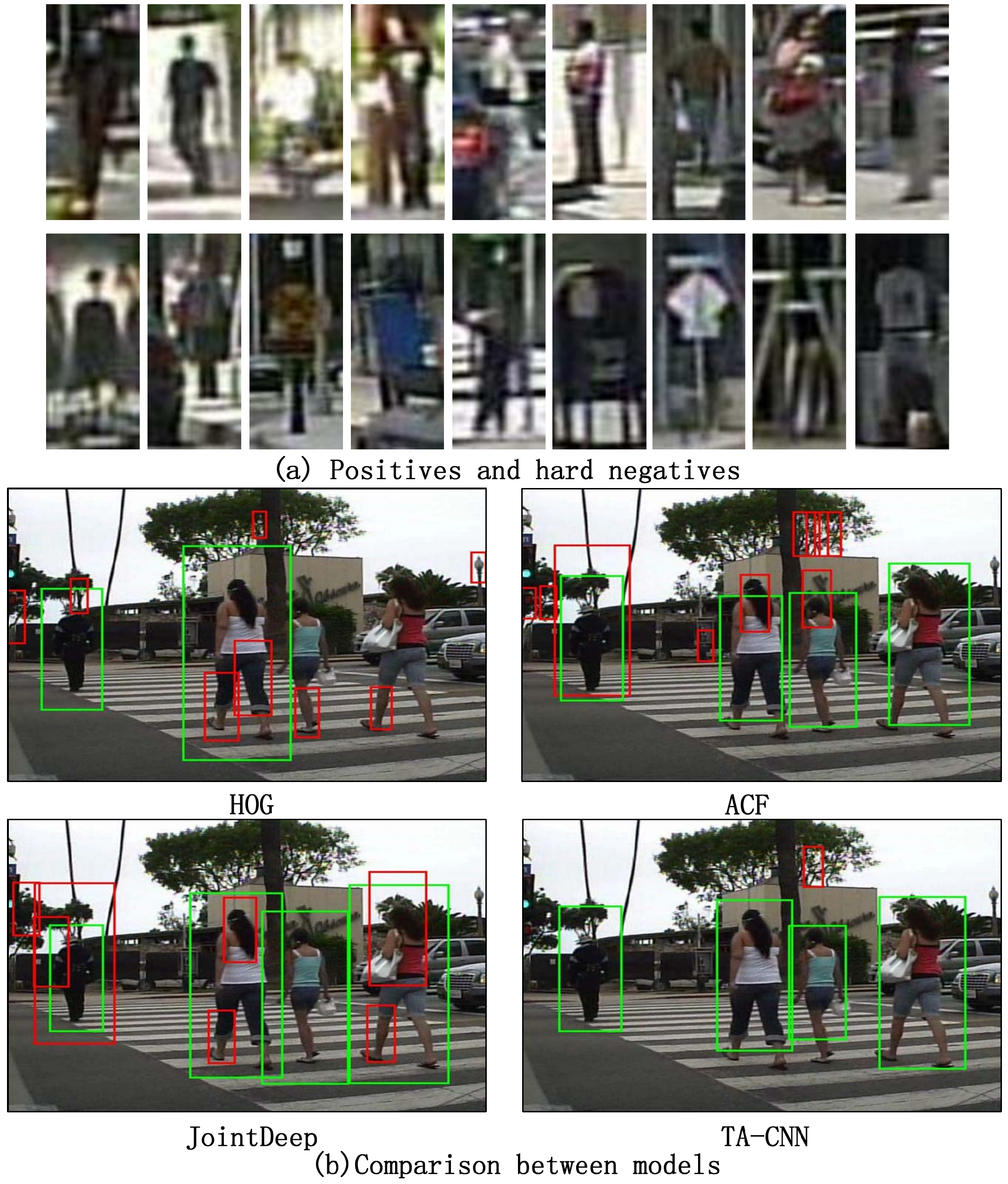}
	\caption{Separating positive samples (pedestrians) from hard negative samples is challenging due to the visual similarity. For example, the first and second row of (a) represent pedestrians and equivocal background samples (hard negatives), respectively. (b) shows that our TA-CNN rejects more hard negatives than detectors using hand-crafted features (HOG \cite{DT05} and ACF \cite{DollarPAMI14pyramids}) and the best-performing deep model (JointDeep \cite{Ouyang2013Joint}).}
	\label{fig:intro}
\end{figure}

Pedestrian detection has attracted broad attentions \cite{DT05,viola2005detecting,sermanet2013,DollarPAMI14pyramids,DollarBMVC09ChnFtrs,Dollar2012PAMI}.
This problem is challenging because of large variation and confusion in human body and background scene, as shown in Fig.\ref{fig:intro} (a), where the positive and hard negative patches have large ambiguity.


Current methods for pedestrian detection can be generally grouped into two categories, the models based on hand-crafted features \cite{viola2005detecting,DT05,DBLP:conf/iccv/WangHY09,DollarBMVC09ChnFtrs,DollarPAMI14pyramids,zhang2013informed,felzenszwalb2010object} and deep models \cite{Ouyang:DBNHuman,Ouyang2013MutualDBN,sermanet2013,Ouyang2013Joint,luo2013switchable}. %
In the first category, conventional methods extracted Haar \cite{viola2005detecting}, HOG\cite{DT05}, or HOG-LBP \cite{DBLP:conf/iccv/WangHY09} from images to train SVM \cite{DT05} or boosting classifiers \cite{DollarBMVC09ChnFtrs}. The learned weights of the classifier (\eg SVM) can be considered as a global template of the entire human body. To account for more complex pose, the hierarchical deformable part models (DPM) \cite{felzenszwalb2010object,zhu2010learning,lin2010shape} learned a mixture of local templates for each body part.
Although they are sufficient to certain pose changes, the feature representations and the classifiers cannot be jointly optimized to improve performance.
In the second category, deep neural networks achieved promising results \cite{Ouyang:DBNHuman,Ouyang2013MutualDBN,sermanet2013,Ouyang2013Joint,luo2013switchable}, 
owing to their capacity to learn middle-level representation.
For example,
%
Ouyang \etal \cite{Ouyang2013Joint} learned features by designing specific hidden layers for the Convolutional Neural Network (CNN), such that features, deformable parts, and pedestrian classification can be jointly optimized.
However, previous deep models treated pedestrian detection as a single binary classification task, they can mainly learn middle-level features, which are not able to capture rich pedestrian variations, as shown in Fig.\ref{fig:intro} (a).

\begin{figure}[t]
	\centering
	\includegraphics[width=1.0\linewidth]{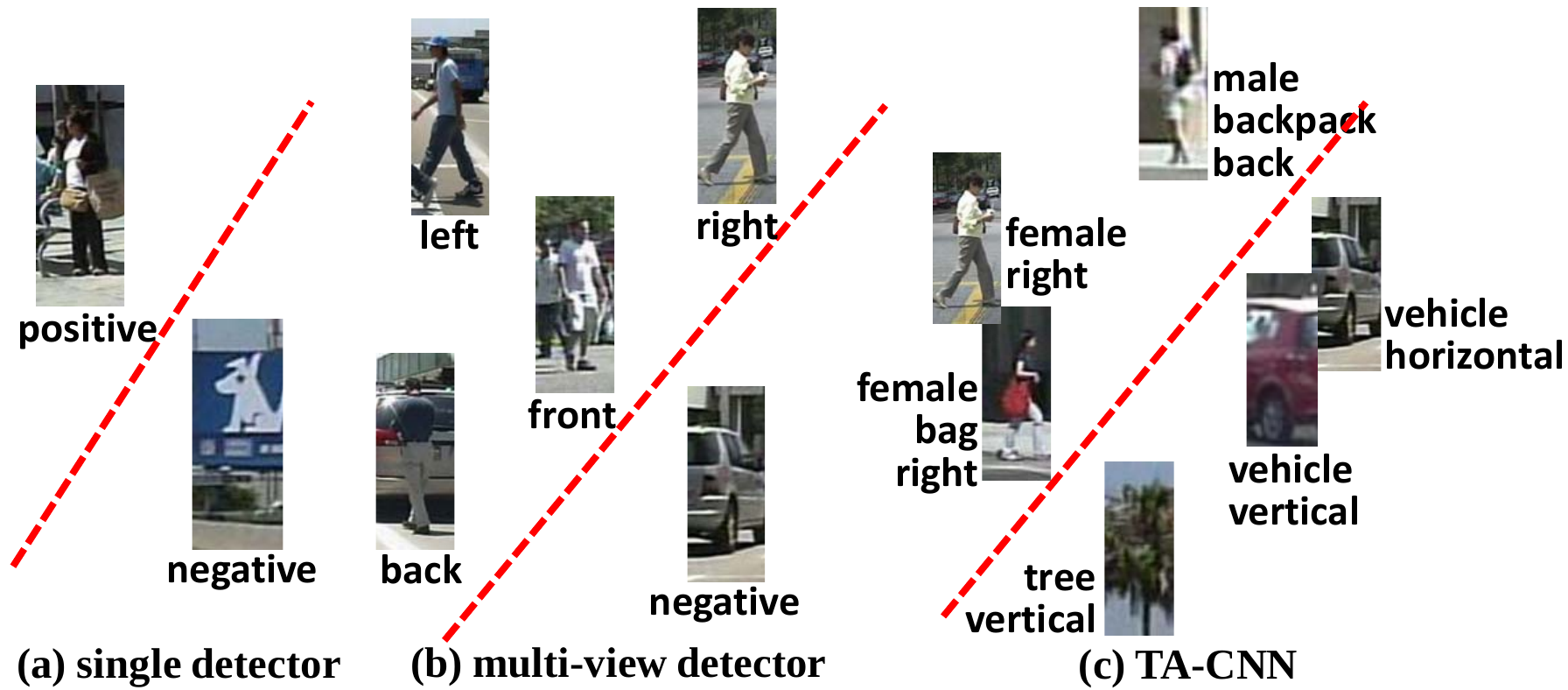}
	\caption{Comparisons between different schemes of pedestrian detectors.}
	\label{fig:multiview}
\end{figure}

To learn high-level representations, this work jointly optimizes pedestrian detection with auxiliary semantic tasks, including pedestrian attributes (\eg `backpack', `gender', and `views') and scene attributes (\eg `vehicle', `tree', and `vertical'). To understand how this work,
we provide an example in Fig.\ref{fig:multiview}. If only a single detector is used to classify all the positive and negative samples in Fig.\ref{fig:multiview} (a), it is difficult to handle complex pedestrian variations.
Therefore, the mixture models of multiple views were developed in Fig.\ref{fig:multiview} (b), \ie pedestrian images in different views are handled by different detectors. If views are treated as one type of semantic tasks, learning pedestrian representation by multiple attributes with deep models actually extends this idea to extreme.
As shown in Fig.\ref{fig:multiview} (c), more supervised information enriches the learned features to account for combinatorial more pedestrian variations. The samples with similar configurations of attributes can be grouped and separated in the high-level feature space.


Specifically, given a pedestrian dataset (denoted by $\mathbf{P}$), the positive image patches are manually labeled with several pedestrian attributes, which are suggested to be valuable for surveillance analysis \cite{handbook}. However, as the number of negatives is significantly larger than the number of positives, we transfer scene attributes information from existing background scene segmentation databases (each one is denoted by $\mathbf{B}$) to the pedestrian dataset, other than annotating them manually.
A novel task-assistant CNN (TA-CNN) is proposed to jointly learn multiple tasks using multiple data sources.
As different $\mathbf{B}$'s may have different data distributions, to reduce these discrepancies, we transfer two types of scene attributes that are carefully chosen, comprising the shared attributes that appear across all the $\mathbf{B}$'s and the unshared attributes that appear in only one of them.
The former one facilitates the learning of shared representation among $\mathbf{B}$'s, whilst the latter one increases diversity of attribute.
Furthermore, to reduce gaps between $\mathbf{P}$ and $\mathbf{B}$'s, we first project each sample in $\mathbf{B}$'s to a structural space of $\mathbf{P}$ and then the projected values are employed as input to train TA-CNN.
Learning TA-CNN is formulated as minimizing a weighted multivariate cross-entropy loss, where both the importance coefficients of tasks and the network parameters can be iteratively solved via stochastic gradient descent \cite{krizhevsky2012imagenet}.

\begin{figure}[t]
	\centering
	\includegraphics[width=1.0\linewidth]{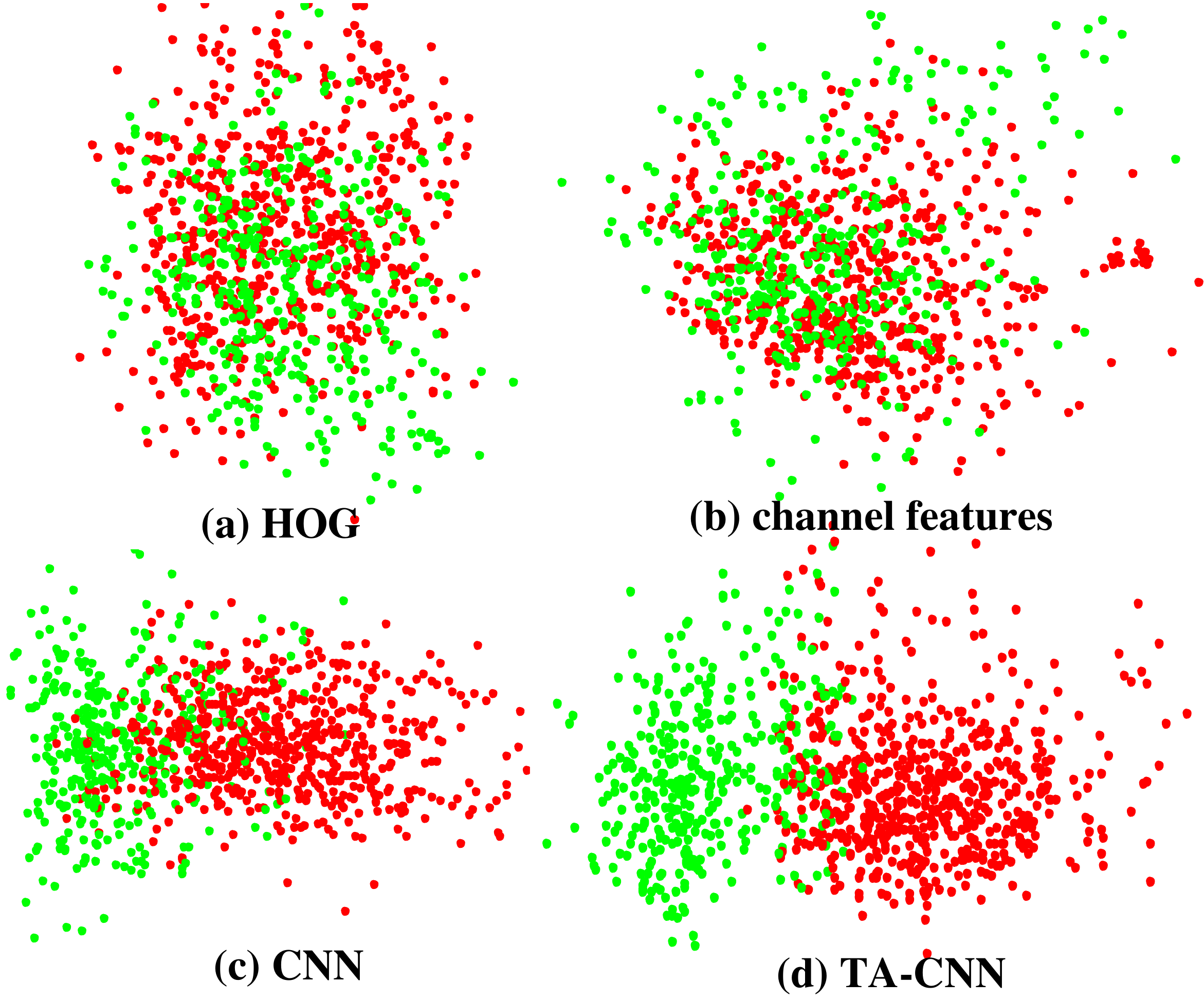}
	\caption{Comparisons of the feature spaces of HOG, channel features, CNN that models pedestrian detection as binary classification, and TA-CNN, using the Caltech-Test set \cite{Dollar2012PAMI}. The positive and hard negative samples are represented by red and green, respectively.}\label{fig:feature}
\end{figure}

This work has the following main \textbf{contributions}. (1) To our knowledge, this is the first attempt to learn high-level representation for pedestrian detection by jointly optimizing it with semantic attributes, including pedestrian attributes and scene attributes. The scene attributes can be transferred from existing scene datasets without annotating manually. (2) These multiple tasks from multiple sources are trained using a single task-assistant CNN (TA-CNN), which is carefully designed to bridge the gaps between different datasets. A weighted multivariate cross-entropy loss is proposed to learn TA-CNN, by iterating among two steps, updating network parameters with tasks' weights fixed and updating weights with network parameters fixed. (3) We systematically investigate the effectiveness of attributes in pedestrian detection. Extensive experiments on both challenging Caltech \cite{Dollar2012PAMI} and ETH \cite{ess2007depth} datasets demonstrate that TA-CNN outperforms state-of-the-art methods. It reduces miss rates of existing deep models on these datasets by $17$ and $5.5$ percent, respectively.

\subsection{Related Works}
We review recent works in two aspects.

\textbf{Models based on Hand-Crafted Features}
The hand-crafted features, such as HOG, LBP, and channel features, achieved great success in pedestrian detection. For example, Wang \etal \cite{DBLP:conf/iccv/WangHY09} utilized the LBP+HOG features to deal with partial occlusion of pedestrian. Chen \etal \cite{chen2013detection} modeled the context information in a multi-order manner. The deformable part models \cite{felzenszwalb2010object} learned mixture of local templates to account for view and pose variations. Moreover, Doll\'{a}r \etal proposed Integral Channel Features (ICF) \cite{DollarBMVC09ChnFtrs} and Aggregated Channel Features (ACF) \cite{DollarPAMI14pyramids}, both of which consist of gradient histogram, gradients, and LUV, and can be efficiently extracted. Benenson \etal \cite{benenson2012pedestrian} combined channel features and  depth information.
However, the representation of hand-crafted features cannot be optimized for pedestrian detection. They are not able to capture large variations, as shown in Fig.\ref{fig:feature} (a) and (b).



\textbf{Deep Models} Deep learning methods can learn features from raw pixels to improve the performance of pedestrian detection.
%
For example, ConvNet \cite{sermanet2013} employed convolutional sparse coding to unsupervised pre-train CNN for pedestrian detection. Ouyang \etal \cite{Ouyang:DBNHuman} jointly learned features and the visibility of different body parts to handle occlusion. The JointDeep model \cite{Ouyang2013Joint} designed a deformation hidden layer for CNN to model mixture poses information.
Unlike the previous deep models that formulated pedestrian detection as a single binary classification task, TA-CNN jointly optimizes pedestrian detection with related semantic tasks, and the learned features are more robust to large variations, as shown in Fig.\ref{fig:feature} (c) and (d).

\begin{figure*}[t]
	\centering
	\includegraphics[width=0.95\textwidth]{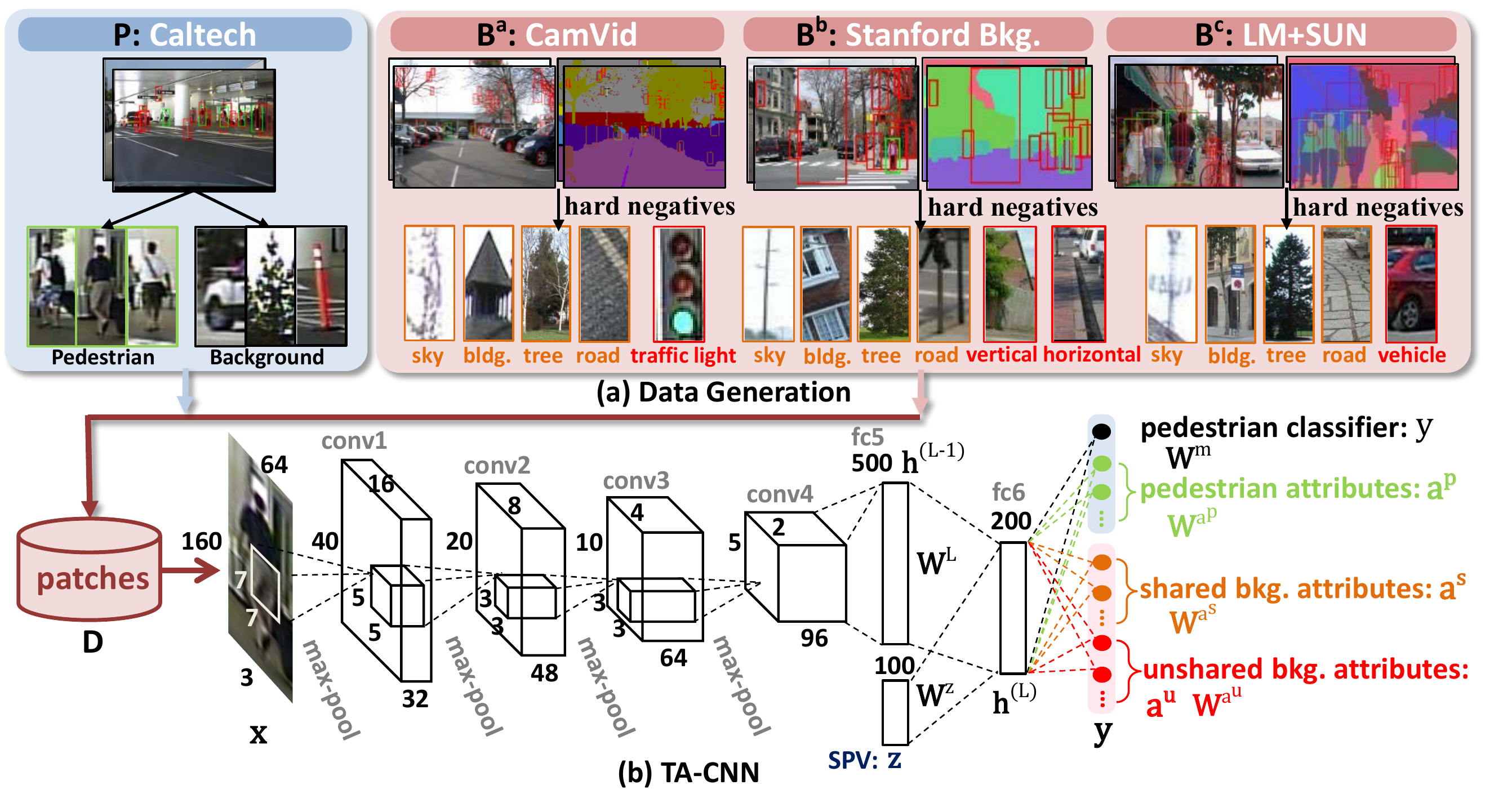}
	\caption{The proposed pipeline for pedestrian detection (\textbf{Best viewed in color}).}
	\label{fig:pipeline}
\end{figure*}

\section{Our Approach}

\begin{figure}[t]
	\centering
	\includegraphics[width=1.0\linewidth]{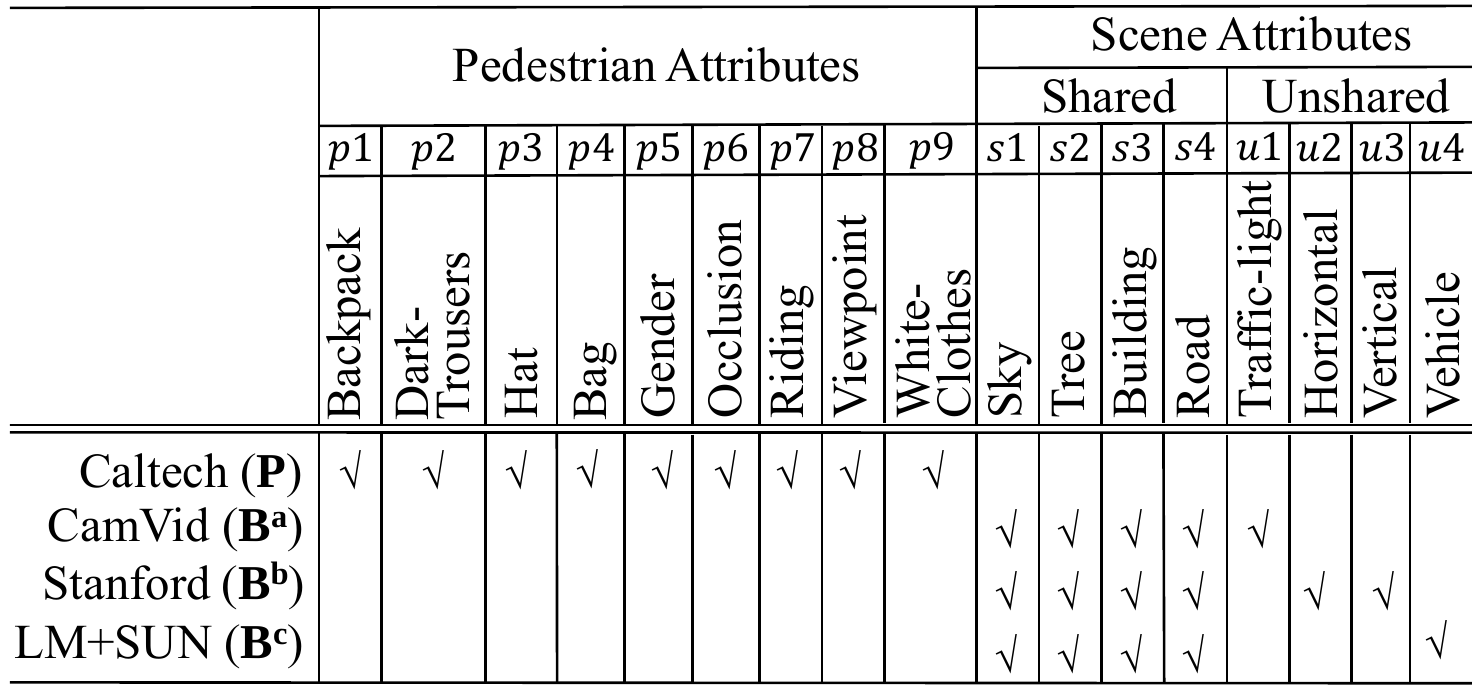}
	\caption{Attribute summarization.}\label{fig:AttributeTabel}
\end{figure}

\textbf{Method Overview} Fig.\ref{fig:pipeline} shows our pipeline of pedestrian detection, where pedestrian classification, pedestrian attributes, and scene attributes are jointly learned by a single TA-CNN. Given a pedestrian dataset $\mathbf{P}$, for example Caltech \cite{Dollar2012PAMI}, we manually label the positive patches with nine pedestrian attributes, which are listed in Fig.\ref{fig:AttributeTabel}. Most of them are suggested by the UK Home Office and UK police to be valuable in surveillance analysis \cite{handbook}.
Since the number of negative patches in $\mathbf{P}$ is significantly larger than the number of positives, we transfer scene attribute information from three public scene segmentation datasets to $\mathbf{P}$, as shown in Fig.\ref{fig:pipeline} (a), including CamVid ($\mathbf{B}^a$) \cite{BrostowSFC:ECCV08}, Stanford Background ($\mathbf{B}^b$) \cite{gould2009decomposing}, and LM+SUN ($\mathbf{B}^c$) \cite{tighe2010superparsing}, where hard negatives are chosen by applying a simple yet fast pedestrian detector \cite{DollarPAMI14pyramids} on these datasets.
As the data in different $\mathbf{B}$'s are sampled from different distributions, we carefully select two types of attributes, the shared attributes (outlined in orange) that present in all $\mathbf{B}$'s and the unshared attributes (outlined in red) that appear only in one of them. This is done because the former one enables the learning of shared representation across $\mathbf{B}$'s, while the latter one enhances diversity of attribute. All chosen attributes are summarized in Fig.\ref{fig:AttributeTabel}, where shows that data from different sources have different subset of attribute labels. For example, pedestrian attributes only present in $\mathbf{P}$, shared attributes present in all $\mathbf{B}$'s, and the unshared attributes present in one of them, \eg 'traffic light' of $\mathbf{B}^a$.

We construct a training set $\mathbf{D}$ by combing patches cropped from both $\mathbf{P}$ and $\mathbf{B}$'s. Let $\mathbf{D}=\{(\bx_n,\by_n)\}_{n=1}^N$ be a set of image patches and their labels, where each $\by_n=(y_n,\bo_n^p,\bo_n^s,\bo_n^u)$ is a four-tuple\footnote{In this paper, scalar variable is denoted by normal letter, while set, vector, or matrix is denoted as boldface letter.}. Specifically, $y_n$ denotes a binary label, indicating whether an image patch is pedestrian or not. $\bo_n^p=\{o_{n}^{pi}\}_{i=1}^9$, $\bo_n^s=\{o_{n}^{si}\}_{i=1}^4$, and $\bo_{n}^u=\{o_{n}^{ui}\}_{i=1}^4$ are three sets of binary labels, representing the pedestrian, shared scene, and unshared scene attributes, respectively.
As shown in Fig.\ref{fig:pipeline} (b), TA-CNN employs image patch $\bx_n$ as input and predicts $\by_n$, by stacking four convolutional layers (conv1 to conv4), four max-pooling layers, and two fully-connected layers (fc5 and fc6). This structure is inspired by the AlexNet \cite{krizhevsky2012imagenet} for large-scale general object categorization. However, as the difficulty of pedestrian detection is different from general object categorization, we remove one convolutional layer of AlexNet and reduce the number of parameters at all remaining layers. The subsequent structure of TA-CNN is specified in Fig.\ref{fig:pipeline} (b).

%

%

\textbf{Formulation of TA-CNN} Each hidden layer of TA-CNN from conv1 to conv4 is computed recursively by convolution and max-pooling, which are formulated as
\begin{eqnarray}
	 \bh_n^{v(l)}&=&\mathrm{relu}(b^{v(l)}+\sum_{u}\bk^{vu(l)}\ast\bh_n^{u(l-1)}),\label{eq:conv}\\
	\bh_{n(i,j)}^{v(l)}&=&\max_{\forall (p,q)\in\Omega_{(i,j)}}\{\bh_{n(p,q)}^{v(l)}\}.\label{eq:max}
\end{eqnarray}
In Eqn.(\ref{eq:conv}), $\mathrm{relu}(x)=\max(0,x)$ is the rectified linear function \cite{nair2010rectified} and $\ast$ denotes the convolution operator applied on every pixel of the feature map $\bh_n^{u(l-1)}$, where $\bh_n^{u(l-1)}$ and $\bh_n^{v(l)}$ stand for the $u$-th input channel at the $l-1$ layer and the $v$-th output channel at the $l$ layer, respectively. $\bk^{vu(l)}$ and $b^{v(l)}$ denote the filters and bias.
In Eqn.(\ref{eq:max}), the feature map $\bh_n^{v(l)}$ is partitioned into grid with overlapping cells, each of which is denoted as $\Omega_{(i,j)}$, where $(i,j)$ indicates the cell index. The max-pooling compares value at each location $(p,q)$ of a cell and outputs the maximum value of each cell.

Each hidden layer in fc5 and fc6 is obtained by
\begin{equation}\label{eq:fc}
	\bh_n^{(l)}=\mathrm{relu}(\trans{\bW^{(l)}}\bh_n^{(l-1)}+\bb^{(l)}),
\end{equation}
where the higher level representation is transformed from lower level with a non-linear mapping. $\bW^{(l)}$ and $\bb^{(l)}$ are the weight matrixes and bias vector at the $l$-th layer.

%


TA-CNN can be formulated as minimizing the log posterior probability with respect to a set of network parameters $\mathcal{W}$
\begin{equation}\label{eq:loss1}
	\mathcal{W}^*=\arg\min_{\mathcal{W}}-\sum_{n=1}^N\log p(y_n,\bo_n^p,\bo_n^s,\bo_n^u|\bx_n;\mathcal{W}),
\end{equation}
where $E=-\sum_{n=1}^N\log p(y_n,\bo_n^p,\bo_n^s,\bo_n^u|\bx_n)$ is a complete loss function regarding the entire training set.
Here, we illustrate that the shared attributes $\bo_n^s$ in Eqn.(\ref{eq:loss1}) are crucial to learn shared representation across multiple scene datasets $\mathbf{B}$'s.

For clarity, we keep only the unshared scene attributes $\bo_n^u$ in the loss function, which then becomes $E=-\sum_{n=1}^N\log p(\bo_n^u|\bx_n)$.
Let $\bx^a$ denote the sample of $\mathbf{B}^a$.
A shared representation can be learned if and only if all the samples share at least one target (attribute).
Since the samples are independent,
%
the loss function can be expanded as $E=-\sum_{i=1}^I\log p(o_{i}^{u1}|\bx_{i}^a)-\sum_{j=1}^J\log p(o_{j}^{u2},o_{j}^{u3}|\bx_j^b)-\sum_{k=1}^K\log p(o_{k}^{u4}|\bx_k^c)$, where $I+J+K=N$, implying that each dataset is only used to optimize its corresponding unshared attribute,
although all the datasets and attributes are trained in a single TA-CNN. For instance, the classification model of $o^{u1}$ is learned by using $\mathbf{B}^a$ without leveraging the existence of the other datasets. In other words, the probability of $p(o^{u1}|\bx^a,\bx^b,\bx^c)=p(o^{u1}|\bx^a)$ because of missing labels.
The above formulation is not sufficient to learn shared features among datasets, especially when the data have large differences.
To bridge multiple scene datasets $\mathbf{B}$'s, we introduce the shared attributes $\bo^s$, the loss function develops into $E=-\sum_{n=1}^N\log p(\bo^s_n,\bo^u_n|\bx_n)$, such that TA-CNN can learn a shared representation across $\mathbf{B}$'s because the samples share common targets $\bo^s$, \ie $p(o^{s1},o^{s2},o^{s3},o^{s4}|\bx^a,\bx^b,\bx^c)$.

Now, we reconsider Eqn.(\ref{eq:loss1}), where the loss function can be decomposed similarly, $E=-\sum_{i=1}^I \log p(\bo^s_i,o^{u1}_{i}|\bx^a_i)-\sum_{j=1}^J\log p(\bo^s_j,o_{j}^{u2},o_{j}^{u3}|\bx^b_j)-\sum_{k=1}^K\log p(\bo^s_k,o_{k}^{u4}|\bx^c_k)-\sum_{\ell=1}^L \log p(y_\ell,\bo^p_\ell|\bx_\ell^p)$, with $I+J+K+L=N$. Even though the discrepancies among $\mathbf{B}$'s can be reduced by $\bo^s$, this decomposition shows that gap remains between datasets $\mathbf{P}$ and $\mathbf{B}$'s. To resolve this issue, we compute the structure projection vectors $\bz_n$ for each sample $\bx_n$, and
Eqn.(\ref{eq:loss1}) turns into
\begin{equation}\label{eq:loss2}
	\mathcal{W}^*=\arg\min_{\mathcal{W}}-\sum_{n=1}^N\log p(y_n,\bo_n^p,\bo_n^s,\bo_n^u|\bx_n,\bz_n;\mathcal{W}).
\end{equation}
For example, the first term of the above decomposition can be written as $p(\bo^s_i,o^{u1}_{i}|\bx^a_i,\bz^a_i)$, where $\bz^a_i$ is attained by projecting the corresponding $\bx^a_i$ in $\mathbf{B}^a$ on the feature space of $\mathbf{P}$. This procedure is explained below.
Here $\bz^a_i$ is used to bridge multiple datasets, because samples from different datasets are projected to a common space of \textbf{P}. TA-CNN adopts a pair of data $(\bx^a_i,\bz^a_i)$ as input (see Fig.\ref{fig:pipeline} (b)). All the remaining terms can be derived in a similar way.

\begin{figure}[t]
	\centering
	\includegraphics[width=1.0\linewidth]{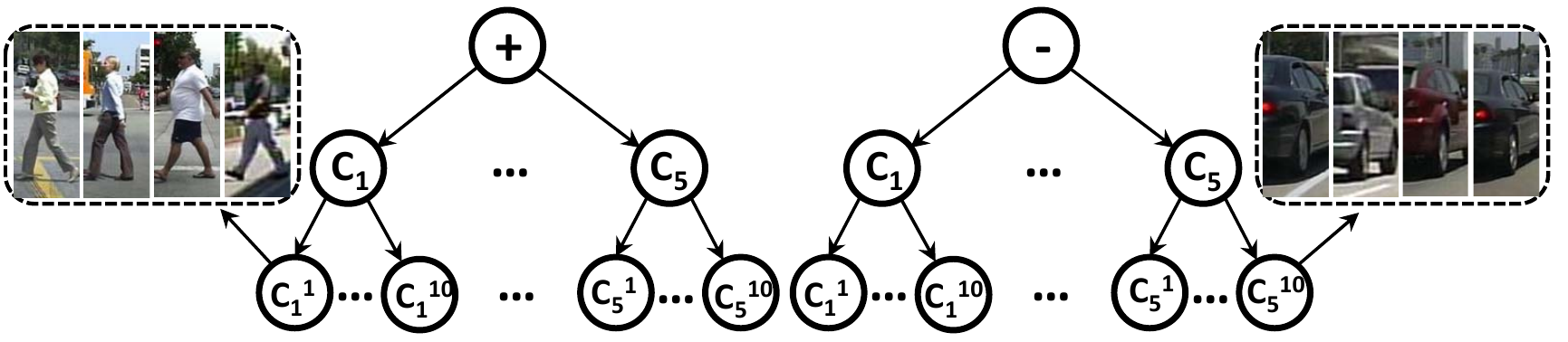}
	\caption{The computation of the structural projection vector (SPV).}
	\label{fig:cluster}
\end{figure}

\textbf{Structure Projection Vector}
As shown in Fig.\ref{fig:cluster}, to close the gap between $\mathbf{P}$ and $\mathbf{B}$'s, we calculate the structure projection vector (SPV) for each sample by organizing the positive (+) and negative (-) data of $\mathbf{P}$ into two tree structures, respectively. Each tree has depth that equals three and partitions the data top-down, where each child node groups the data of its parent into clusters, for example $C^1_1$ and $C^{10}_5$. Then, SPV of each sample is obtained by concatenating the distance between it and the mean of each leaf node. Specifically, at each parent node, we extract HOG feature for each sample and apply k-means to group the data. We partition the data into five clusters ($C_1$ to $C_5$) in the first level, and then each of them is further partitioned into ten clusters, \eg $C_1^1$ to $C_1^{10}$.

%
%
%
%
%

\section{Learning Task-Assistant CNN}

To learn network parameters $\mathcal{W}$, a natural way is to reformulate Eqn.(\ref{eq:loss2}) as the softmax loss functions similar to the previous methods. We have\footnote{We drop the sample index $n$ in the remaining derivation for clarity.}
\begin{equation}\label{eq:loss3}
	\begin{split}
		E\triangleq&-y\log p(y|\bx,\bz)-\sum_{i=1}^9\alpha_i o^{pi}\log p(o^{pi}|\bx,\bz)\\ &-\sum_{j=1}^4\beta_j o^{sj}\log p(o^{sj}|\bx,\bz)-\sum_{k=1}^4\gamma_k o^{uk}\log p(o^{uk}|\bx,\bz),
	\end{split}
\end{equation}
where the \emph{main task} is to predict the pedestrian label $y$ and the attribute estimations, \ie $o^{pi}$, $o^{sj}$, and $o^{uk}$, are auxiliary semantic tasks. $\alpha$, $\beta$, and $\gamma$ denote the importance coefficients to associate multiple tasks. Here, $p(y|\bx,\bz)$, $p(o^{pi}|\bx,\bz)$, $p(o^{sj}|\bx,\bz)$, and $p(o^{uk}|\bx,\bz)$ are modeled by softmax functions, for example, $p(y=0|\bx,\bz)=\frac{\exp(\trans{\bW^{m}_{\cdot1}}\bh^{(L)})}{\exp(\trans{\bW^{m}_{\cdot1}}\bh^{(L)})+\exp(\trans{\bW^{m}_{\cdot2}}\bh^{(L)})}$, where $\bh^{(L)}$ and $\bW^{m}$ indicate the top-layer feature vector and the parameter matrix of the main task $y$ respectively, as shown in Fig.\ref{fig:pipeline} (b), and $\bh^{(L)}$ is obtained by $\bh^{(L)}=\mathrm{relu}(\bW^{(L)}\bh^{(L-1)}+\bb^{(L)}+\bW^z\bz+\bb^z)$.

Eqn.(\ref{eq:loss3}) optimizes eighteen loss functions together. It has two main drawbacks.
First, since different tasks have different convergence rates, training many tasks together suffers from over-fitting. Previous works prevented over-fitting by adjusting the importance coefficients. However, they are determined in a heuristic manner, such as early stopping \cite{zhang2014multitask}, other than estimating in the learning procedure. Second, if the dimension of the features $\bh^{(L)}$ is high, the number of parameters at the top-layer increases exponentially. For example, if the feature vector $\bh^{(L)}$ has $H$ dimensions, the weight matrix of each two-state variable (\eg $\bW^m$ of the main task) has $2\times H$ parameters, whilst the weight matrix of the four-state variable `viewpoint' has $4\times H$ parameters\footnote{All tasks are binary classification (\ie two states) except the pedestrian attribute `viewpoint', which has four states, including `front', `back', `left', and `right'.}. As we have seventeen two-state variables and one four-state variable, the total number of parameters at the top-layer is $17\times2\times H+4\times H=38H$.

To resolve the above issues, we cast learning multiple tasks in Eqn.(\ref{eq:loss3}) as optimizing a single weighted multivariate cross-entropy loss, which can not only learn a compact weight matrix but also iteratively estimate the importance coefficients,
\begin{equation}\label{eq:loss4}
	\begin{split}
		E\triangleq&-\trans{\by}\mathrm{diag}(\bla)\log p(\by|\bx,\bz)\\
		 &-\trans{(\textbf{1}-\by)}\mathrm{diag}(\bla)(\log\textbf{1}-p(\by|\bx,\bz),
	\end{split}
\end{equation}
where $\bla$ denotes a vector of importance coefficients and $\mathrm{diag}(\cdot)$ represents a diagonal matrix. Here, $\by=(y,\bo^p,\bo^s,\bo^u)$ is a vector of binary labels, concatenating the pedestrian label and all attribute labels. Note that each two-state (four-state) variable can be described by one bit (two bits). Since we have seventeen two-state variables and one four-state variable, the weight matrix at the top layer, denoted as $\bW^y$ in this case, has $17\times H+2\times H=19H$ parameters, which reduces the number of parameters by half, \ie $19H$ compared to $38H$ of Eqn.(\ref{eq:loss3}).
Moreover, $p(\by|\bx,\bz)$ is modeled by sigmoid function, \ie $p(\by|\bx,\bz)=\frac{1}{1+\exp(-\trans{\bW^y}\bh^{(L)})}$, where $\bh^{(L)}$ is achieved in the same way as in Eqn.(\ref{eq:loss3}).

The optimization of Eqn.(\ref{eq:loss4}) iterates between two steps, updating network parameters with the importance coefficients fixed and updating coefficients with the network parameters fixed.

\textbf{Learning Network Parameters} The network parameters are updated by minimizing Eqn.(\ref{eq:loss4}) using stochastic gradient descent \cite{krizhevsky2012imagenet} and back-propagation (BP) \cite{rumelhart1986learning}, where the error of the output layer is propagated top-down to update filters or weights at each layer.
For example, the weight matrix of the $L$-th layer in the $t+1$-th iteration, $\bW_{t+1}^{y}$, is attained by
\begin{equation}\label{eq:w}
\begin{split}
&\bW_{t+1}^{y}=\bW_t^{y}+\Delta_{t+1},\\
&\Delta_{t+1}=0.9\cdot\Delta_t-0.001\cdot\epsilon\cdot \bW_t^{y}-\epsilon\cdot\frac{\partial E}{\partial \bW_t^{y}}.
\end{split}
\end{equation}
Here, $t$ is the index of training iteration. $\Delta$ is the momentum variable, $\epsilon$ is the learning rate, and $\frac{\partial E}{\partial \bW_t^{y}}=\bh^{(L)}\trans{e^{(L)}}$ is the derivative calculated by the outer product of the back-propagation error $e^{(L)}$ and the hidden features $\bh^{(L)}$.
The BP procedure is similar to \cite{krizhevsky2012imagenet}. The main difference is how to compute error at the $L$-th layer.
%
%
In the traditional BP algorithm, the error $e^{(L)}$ at the $L$-th layer is obtained by the gradient of Eqn.(\ref{eq:loss4}), indicating the loss, \ie $e^{(L)}=\overline{\by}-\by$, where $\overline{\by}$ denotes the predicted labels.
%
%
However, unlike the conventional BP where all the labels are observed, each of our dataset only covers a subset of attributes. Let $\widehat{\bo}$ signify the unobserved labels. The posterior probability of Eqn.(\ref{eq:loss4}) becomes $p(\by_{\setminus\widehat{\bo}},\widehat{\bo}|\bx,\bz)$,
where $\by_{\setminus\widehat{\bo}}$ specifies the labels $\by$ excluding $\widehat{\bo}$.
Here we demonstrate that $\widehat{\bo}$ can be simply marginalized out, since the labels are independent. We have $\sum_{\widehat{\bo}}p(\by_{\setminus\widehat{\bo}},\widehat{\bo}|\bx,\bz)=$ $p(\by_{\setminus\widehat{\bo}}|\bx,\bz)\cdot\sum_{\widehat{o_1}}p(\widehat{o_1}|\bx,\bz)\cdot\sum_{\widehat{o_2}}p(\widehat{o_2}|\bx,\bz)\cdot...\cdot\sum_{\widehat{o_j}}p(\widehat{o_j}|\bx,\bz)=p(\by_{\setminus\widehat{\bo}}|\bx,\bz)$.
Therefore, the error $e^{(L)}$ of Eqn.(\ref{eq:loss4}) can be computed as
\begin{equation}\label{eq:error1}
	e^{(L)}=
	\begin{cases}
		\overline{y}-y, & \text{if $y\in\by_{\setminus\widehat{\bo}}$,}\\
		0,  & \text{otherwise,}
	\end{cases}
\end{equation}
which demonstrates that the errors of the missing labels will not be propagated no matter whether their predictions are correct or not.
%



\textbf{Learning Importance Coefficients} We update the importance coefficients with the network parameters fixed, by minimizing the posterior probability $p(\bla|\bx,\by)=\frac{p(\bx,\by|\bla)p(\bla)}{p(\bx,\by)}$ as introduced in \cite{caruana1998multitask}. Taking the negative logarithm of the posterior, the problem develops into
\begin{equation}\label{eq:coef}
	\begin{split}
		\arg\min_{\bla}-\log p(\bx,\by|\bla)-\log p(\bla)+\log p(\bx,\by),
	\end{split}
\end{equation}
where the first term, $\log p(\bx,\by|\bla)$, is a log likelihood similar to Eqn.(\ref{eq:loss4}), measuring the evidence of selecting importance coefficients $\bla$.
The second term specifies a log prior of $\bla$. To avoid trivial solution, \ie exists $\lambda_i\in\bla$ equals zero, we have $\log p(\bla)=\sum_{i=1}-\frac{1}{\sigma^2}\|\lambda_i-1\|_2^2$, showing that each coefficient is regularized by a Gaussian prior with mean `$1$' and standard deviation $\sigma$.
This implies that each $\lambda_i\in\bla$ should not deviate too much from one, because we assume all tasks have equal contributions at the very beginning.
Let $\lambda_1$ be the coefficient of the main task. We fix $\lambda_1=1$ through out the learning procedure, as our goal is to optimize the main task with the help of the auxiliary tasks.
The third term is a normalization constant,
which can be simply modeled as a constant scalar. In this work, we adopted the restricted Boltzmann machine (RBM) \cite{hinton2006reducing} to learn $p(\bx,\by)$, because RBM can well model the data space. In other words, we can measure the predictions of the coefficients with respect to the importance of each sample. Note that RBM can be learned off-line and $p(\bx,\by)$ can be stored in a probability table for fast indexing.

\begin{algorithm}[t]
	\caption{Learning TA-CNN}\label{alg:1}
	\KwData{Training set $\mathbf{D}=\{(\bx_n,\by_n)\}_{n=1}^N$;}
	\KwResult{Network parameters $\mathcal{W}$ and importance coefficients $\bla$;}
	\BlankLine
	Train RBM of $p(\bx,\by)$, and calculate and store the probability table of $p(\bx,\by)$\;
	\While{not stopping criterion}{
		1.~update $\mathcal{W}$ with $\bla$ fixed:~repeat the below process until a local minima is reached,\\
		\For{a minibatch of $\bx_n$}{
			forward propagation by using Eqn.(\ref{eq:conv}), (\ref{eq:max}), and (\ref{eq:fc})\;
			backward propagation to update network filters and weights by BP\;
		}
		2.~update $\bla$ with $\mathcal{W}$ fixed by solving Eqn.(\ref{eq:coef})\;
	}
\end{algorithm}

Intuitively, coefficient learning is similar to the process below. At the very beginning, all the tasks have equal importance. In the training stage, for those tasks whose values of the loss function are stable but large, we decrease their weights, because they may not relate to the main task or begin to over-fit the data. However, we penalize the coefficient that is approaching zero, preventing the corresponding task from suspension. For those tasks have small values of loss, their weights could be increased, since these tasks are highly related to the main task, \ie whose error rates are synchronously decreased with the main task.
In practice, all the tasks' coefficients in our experiments become $0.1\sim0.2$ when training converges, except the main task whose weight is fixed and equals one. Learning of TA-CNN is summarized in Algorithm \ref{alg:1}. Typically, we run the first step for sufficient number of iterations to reach a local minima, and then perform the second step to update the coefficients. This strategy can help avoid getting stuck at local minima.

Here, we explain the third term in details. With the RBM, we have
$$\log p(\bx,\by)=\log\sum_\bh\exp\Big(-E(\bx,\by,\bh)\Big),$$
which represents the free energy \cite{hinton2006reducing} of RBM. Specifically, $E(\bx,\by,\bh)=-\trans{\bx}\bW^{xh}\bh-\trans{\bx}\bb^x-\trans{\by}\bW^{yh}\bh-\trans{\by}\bb^y-\trans{\bh}\bb^h$ is the energy function, which learns the latent binary representation $\bh$ that models the shared hidden space of $\bx$ and $\by$. $\bW^{xh}$ and $\bW^{yh}$ are the projection matrixes capturing the relations between $\bx$ and $\bh$, and $\by$ and $\bh$, respectively, while $\bb^x,\bb^y$, and $\bb^h$ are the biases. The RBM can be solved by the contrastive divergence \cite{hinton2006reducing}. Since the latent variables $\bh$ are independent given $\bx$ and $\by$, $\log p(\bx,\by)$ can be rewritten by integrating over $\bh$, \ie $\log p(\bx,\by)=\sum_{i}\log\Big(1+\exp(b_i^h+\trans{\bx}\bW^{xh}_{\cdot i}+\trans{\by}\bW^{yh}_{\cdot i})\Big)+\trans{\bx}\bb^x+\trans{\by}\bb^y$.
Combining all the above definitions, Eqn.(\ref{eq:coef}) is an unconstrained optimization problem, where the importance coefficients can be efficiently updated by using the L-BFGS algorithm \cite{andrew2007scalable}.




\section{Experiments}
The proposed TA-CNN is evaluated on the Caltech-Test \cite{Dollar2012PAMI} and ETH datasets \cite{ess2007depth}.
We strictly follow the evaluation protocol proposed in \cite{Dollar2012PAMI}, which measures the log average miss rate over nine points ranging from $10^{-2}$ to $10^{0}$ False-Positive-Per-Image.
We compare TA-CNN with the best-performing methods as suggested by the Caltech and ETH benchmarks\footnote{\myurl{http://www.vision.caltech.edu/Image_Datasets/CaltechPedestrians/}} on the \emph{reasonable} subsets, where pedestrians are larger than $49$ pixels height and have $65$ percent visible body parts.

\subsection{Effectiveness of TA-CNN}
We systematically study the effectiveness of TA-CNN in four aspects as follows. In this section, TA-CNN is trained on Caltech-Train and tested on Caltech-Test.

\textbf{Effectiveness of Hard Negative Mining}
To save computational cost, We employ ACF \cite{DollarPAMI14pyramids} for mining hard negatives at the training stage and pruning candidate windows at the testing stage. Two main adjustments are made in ACF. First, we compute the exact feature pyramids at each scale instead of making an estimated aggregation. Second, we increase the number of weak classifiers to enhance the recognition ability.
Afterwards, a higher recall rate is achieved by ACF and it obtains $37.31$ percent miss rate on Caltech-Test.
Then the TA-CNN with only the main task (pedestrian classification) achieved 31.45 percent miss rate by cascading on ACF, obtaining more than $5$ percent improvement.


\begin{table}
	\footnotesize
	\begin{center}
		\setlength{\tabcolsep}{.08em}
		 \begin{tabular*}{1.0\linewidth}{@{\extracolsep{\fill}}c|c|c|c|c|c|c|c|c|c|c}
			\hline
			\rotatebox{90}{\textbf{main task}} & \rotatebox{90}{backpack} & \rotatebox{90}{dark-trousers} & \rotatebox{90}{hat} & \rotatebox{90}{bag} & \rotatebox{90}{gender} & \rotatebox{90}{occlusion} & \rotatebox{90}{riding} & \rotatebox{90}{white-cloth} & \rotatebox{90}{viewpoint} & \rotatebox{90}{\textbf{All}} \\
			\hline
			\hline
			 $31.45$ & $30.44$ & $29.83$ & $28.89$ & $30.77$ & $30.70$& $29.36$ & $28.83$& $30.22$& $28.20$& $\textbf{25.64}$\\
			\hline
		\end{tabular*}
	\end{center}
	\caption{Log-average miss rate (\%) on Caltech-Test with pedestrian attribute learning tasks.}
	\label{tab:posattribute}
\end{table}%

\begin{table}
	\footnotesize
	\begin{center}
		\setlength{\tabcolsep}{.12em}
		 \begin{tabular*}{1.0\linewidth}{@{\extracolsep{\fill}}c|c|c|c|c|c|c|c|c|c}
			\hline
			& \rotatebox{90}{\textbf{main task}} & \rotatebox{90}{sky}& \rotatebox{90}{tree} & \rotatebox{90}{building} & \rotatebox{90}{road} & \rotatebox{90}{vehicle} & \rotatebox{90}{traffic-light} & \rotatebox{90}{vertical} & \rotatebox{90}{horizontal} \\
			\hline
			\hline
			Neg. & \multirow{2}{*}{$31.45$}& $31.07$ & $30.92$ & $31.16$ & $31.02$ & $30.75$ & $30.85$ & $30.91$ & $30.96$ \\
			Attr. & & $30.79$& $30.50$ & $30.90$ & $30.54$ & $29.41$ & $28.92$ &$30.03$& $30.40$\\
			\hline
		\end{tabular*}
	\end{center}
	\caption{Log-average miss rate (\%) on Caltech-Test with scene attribute learning tasks.}
	\label{tab:negattribute}
\end{table}%

\textbf{Effectiveness of Pedestrian Attributes}
We investigate how different pedestrian attributes can help improve the main task. To this end, we train TA-CNN by combing the main task with each of the pedestrian attributes, and the miss rates are reported in Table \ref{tab:posattribute}, where shows that `viewpoint' is the most effective attribute, which improves the miss rate by $3.25$ percent, because `viewpoint' captures the global information of pedestrian. The attribute capture the pose information also attains significant improvement, \eg $2.62$ percent by `riding'. Interestingly, among those attributes modeling local information, `hat' performs the best, reducing the miss rate by $2.56$ percent. We observe that this result is consistent with previous works, SpatialPooling \cite{paisitkriangkrai2014strengthening} and InformedHaar \cite{zhang2013informed}, which showed that head is the most informative body parts for pedestrian detection. When combining all the pedestrian attributes, TA-CNN achieved 25.64 percent miss rate, improving the main task by $6$ percent.


\textbf{Effectiveness of Scene Attributes}
Similarly, we study how different scene attributes can improve pedestrian detection. We train TA-CNN combining the main task with each scene attribute. For each attribute, we select $5,000$ hard negative samples from its corresponding dataset. For example, we crop five thousand patches for `vertical' from the Stanford Background dataset. We test two settings, denoted as ``Neg.'' and ``Attr.''. In the first setting, we label the five thousand patches as negative samples. In the second setting, these patches are assigned to their original attribute labels. The former one uses more negative samples compared to the TA-CNN (main task), whilst the latter one employs attribute information.

The results are reported in Table \ref{tab:negattribute}, where shows that `traffic-light' improves the main task by $2.53$ percent, revealing that the patches of `traffic-light' are easily confused with positives. This is consistent when we exam the hard negative samples of most of the pedestrian detectors.
Besides, the `vertical' background patches are more effective than the `horizontal' background patches, corresponding to the fact that hard negative patches are more likely to present vertically.

\textbf{Attribute Prediction}
We also consider the accuracy of attribute prediction and find that the averaged accuracy of all the attributes exceeds $75$ percent. We select the pedestrian attribute `viewpoint' as illustration. In Table \ref{viewpoint}, we report the confusion matrix of `viewpoint', where the number of detected pedestrians of `front', `'back', `'left', and `right' are $283$, $276$, $220$, $156$ respectively. We observed that `front' and `back' information are relatively easy to capture, rather than the `left' and `right', which are more likely to confuse with each other, \eg $21+40=61$ mis-classified samples.

\begin{table}
	\footnotesize
	\begin{center}
		\begin{tabular}{c|c|c|c|c|c}
			\hline
			\multicolumn{2}{c|}{}&\multicolumn{4}{|c}{Predict State}\\\cline{3-6}
			\multicolumn{2}{c|}{}& Frontal & Back & Left & Right \\
			\hline
			& Frontal & $226$ & $32$ & $15$ & $10$\\\cline{2-6}
			True & Back & $24$ & $232$ & $12$ & $8$\\\cline{2-6}
			State & Left & $22$ & $13$ & $164$ & $21$\\\cline{2-6}
			& Right & $5$ & $15$ & $40$ & $96$\\\cline{2-6}
			\hline
			\multicolumn{2}{c|}{Accuracy}& $0.816$ & $0.796$ & $0.701$ & $0.711$\\
			\hline
		\end{tabular}
	\end{center}
	\caption{View-point estimation results on Caltech-Test.}\label{viewpoint}
\end{table}


\subsection{Overall Performance on Caltech}\label{sec:caltech}
\begin{figure}
	\begin{subfigure}{\linewidth}
		\centering
		\includegraphics[width=0.9\linewidth]{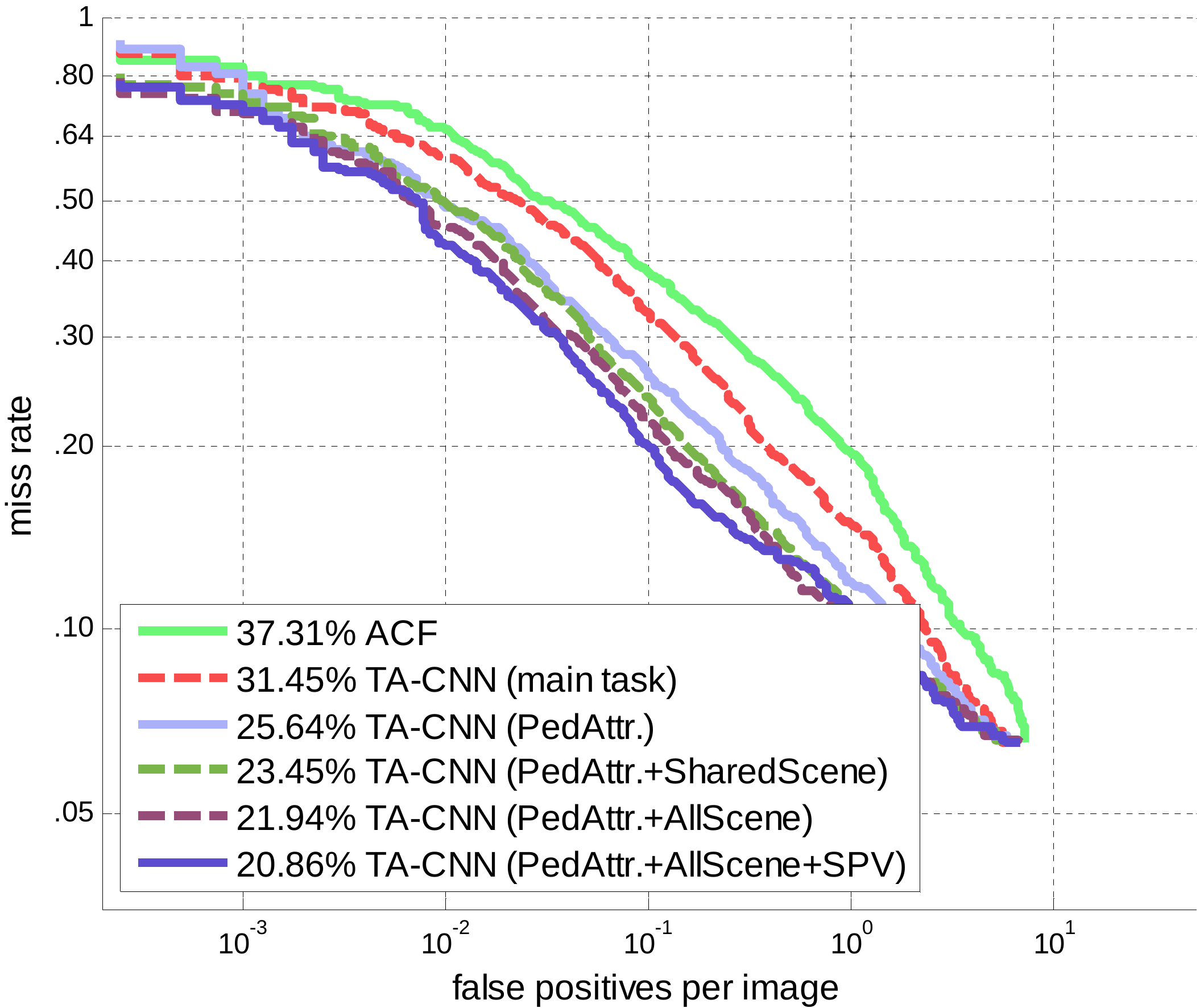}
		\caption{Log-average miss rate reduction procedure}\label{1a}
	\end{subfigure}%
	\\
	\begin{subfigure}{\linewidth}
		\centering
		\includegraphics[width=0.9\linewidth]{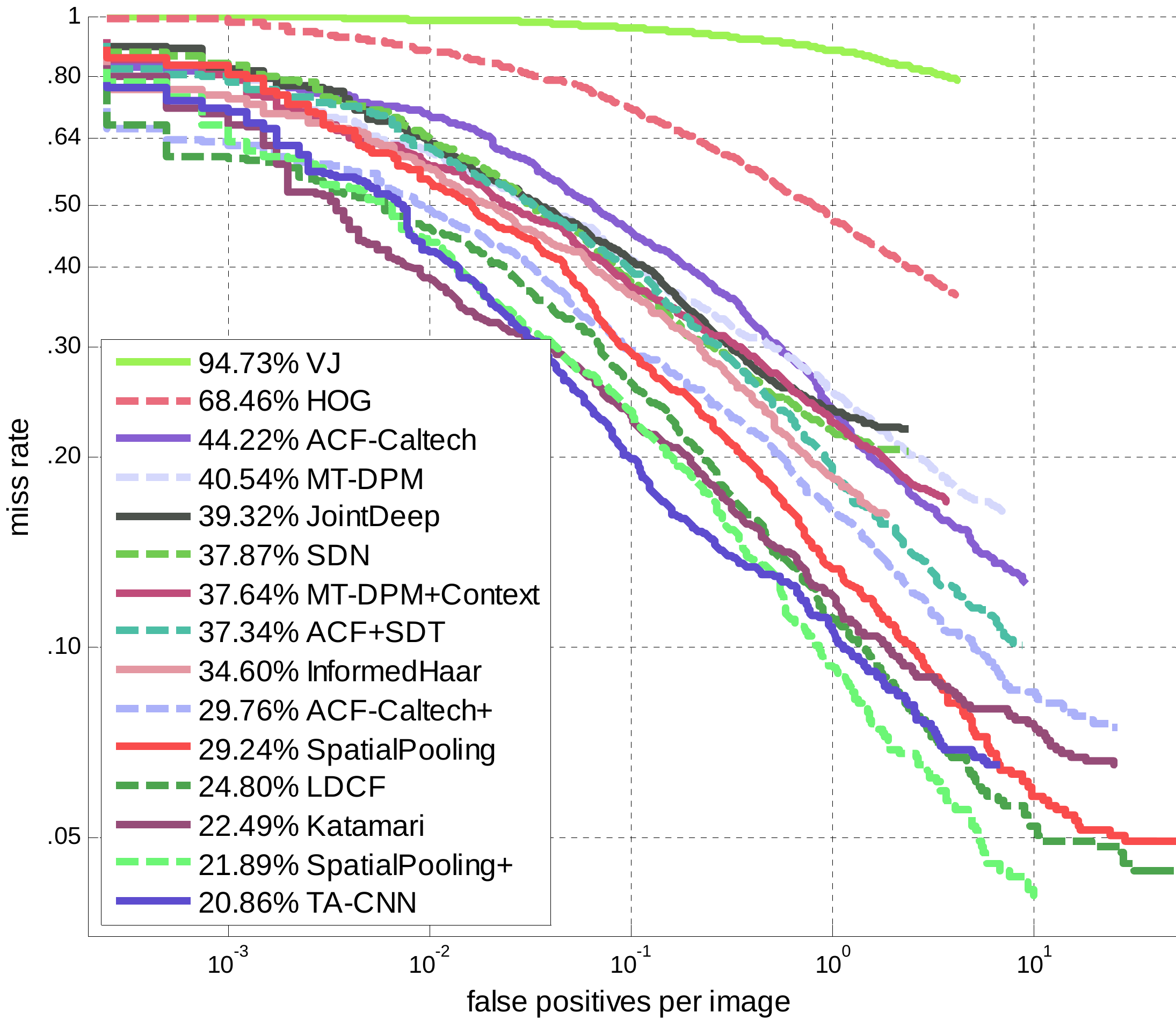}
		\caption{Overall Performance on Caltech-Test}\label{1b}
	\end{subfigure}
	\caption{Results under standard evaluation settings}\label{fig:1}
\end{figure}

We report overall results in two parts. All the results of TA-CNN are obtained by training on Caltech-Train and evaluating on Caltech-Test.
In the first part, we analyze the performance of different components of TA-CNN. As shown in Fig.\ref{1a}, the performances show clear increasing patterns when gradually adding more components. For example, TA-CNN (main task) cascades on ACF and reduces the miss rate of it by more than 5 percent. TA-CNN (PedAttr.+SharedScene) reduces the result of TA-CNN (PedAttr.) by $2.2$ percent, because it can bridge the gaps among multiple scene datasets. After modeling the unshared attributes, the miss rate is further decreased by $1.5$ percent, since more attribute information is incorporated. The final result of $20.86$ miss rate is obtained by using the structure projection vector as input to TA-CNN. Its effectiveness has been demonstrated in Fig.\ref{1a}.

In the second part, we compare the result of TA-CNN with all existing best-performing methods, including VJ \cite{viola2004robust}, HOG \cite{DT05}, ACF-Caltech \cite{DollarPAMI14pyramids}, MT-DPM \cite{yan2013robust}, MT-DPM+Context \cite{yan2013robust}, JointDeep \cite{Ouyang2013Joint}, SDN \cite{luo2013switchable}, ACF+SDT \cite{park2013exploring}, InformedHaar \cite{zhang2013informed}, ACF-Caltech+ \cite{namlocal}, SpatialPooling \cite{paisitkriangkrai2014strengthening}, LDCF \cite{namlocal}, Katamari \cite{benenson2014workshop}, SpatialPooling+ \cite{paisitkriangkrai2014pedestrian}. These works used various features, classifiers, deep networks, and motion and context information. We summarize them as below. Note that TA-CNN dose not employ motion and context information.

\textbf{Features}: Haar (VJ), HOG (HOG, MT-DPM),
Channel-Feature (ACF+Caltech, LDCF); \textbf{Classifiers}:
latent-SVM (MT-DPM), boosting (VJ, ACF+Caltech, SpatialPooling); \textbf{Deep Models}: JointDeep, SDN; \textbf{Motion and context}: MT-DPM+Context, ACF+SDT, Katamari, SpatialPooling+.

Fig.\ref{1b} reports the results. TA-CNN achieved the smallest miss rate compared to all existing methods. Although it only outperforms the second best method (SpatialPooling+ \cite{paisitkriangkrai2014pedestrian}) by $1$ percent, it learns $200$ dimensions high-level features with attributes, other than combining LBP, covariance features, channel features, and video motion as in \cite{paisitkriangkrai2014pedestrian}. Also, the Katamari \cite{benenson2014workshop} method integrates multiple types of features and context information.

%

\begin{figure}[t]
	\centering \includegraphics[width=1.0\linewidth]{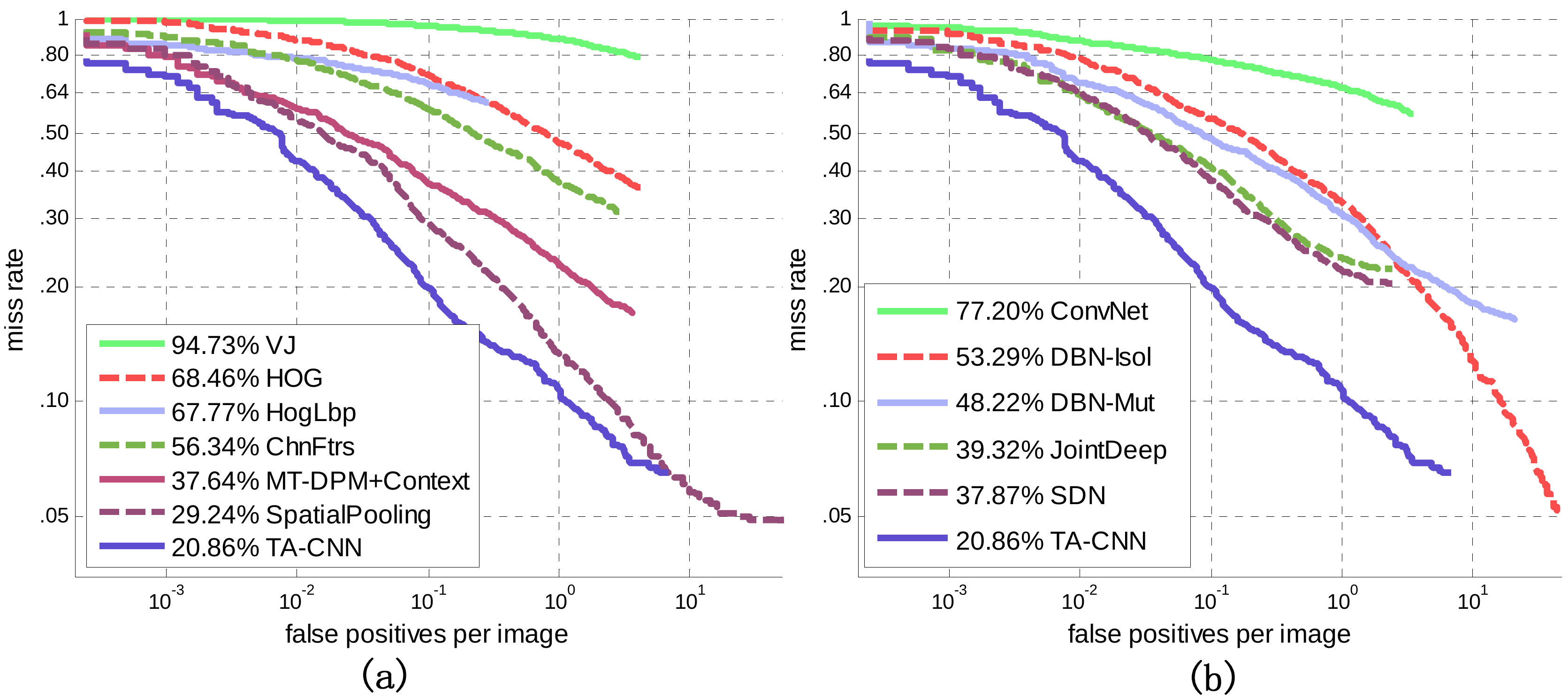}
	\caption{Results on Caltech-Test: (a) comparison with hand-crafted feature based models; (b) comparison with other deep models}
	\label{fig:Comparison}
\end{figure}

\textbf{Hand-crafted Features}
The learned high-level representation of TA-CNN outperforms the conventional hand-crafted features by a large margin, including Haar, HOG, HOG+LBP, and channel features, shown in Fig.\ref{fig:Comparison} (a). For example, it reduced the miss rate by 16 and 9 percent compared to DPM+Context and Spatial Pooling, respectively. DPM+Context combined HOG feature with pose mixture and context information, while SpatialPooling combined multiple features, such as LBP, covariance, and channel features.


\textbf{Deep Models} Fig.\ref{fig:Comparison} (b) shows that TA-CNN surpasses other deep models. For example, TA-CNN outperforms two state-of-the-art deep models, JointDeep and SDN, by 18 and 17  percent, respectively. Both SDN and JointDeep treated pedestrian detection as a single task and thus cannot learn high-level representation to deal with the challenging hard negative samples.

\textbf{Time Complexity}
Training TA-CNN on Caltech-Train with a single GPU takes $3$ hours. At the testing stage, the running time of hard negative mining is $10$ frames per second (FPS) on Matlab with CPU, while TA-CNN runs at $100$ FPS on GPU. In summary, the entire system detects pedestrians from raw $640\times480$ images at around $5$ FPS. The bottleneck is the step of hard negative mining. We consider to migrate it to GPU platform.

\subsection{Overall Performance on ETH}

\begin{figure}[t]
	\centering
	\includegraphics[width=0.9\linewidth]{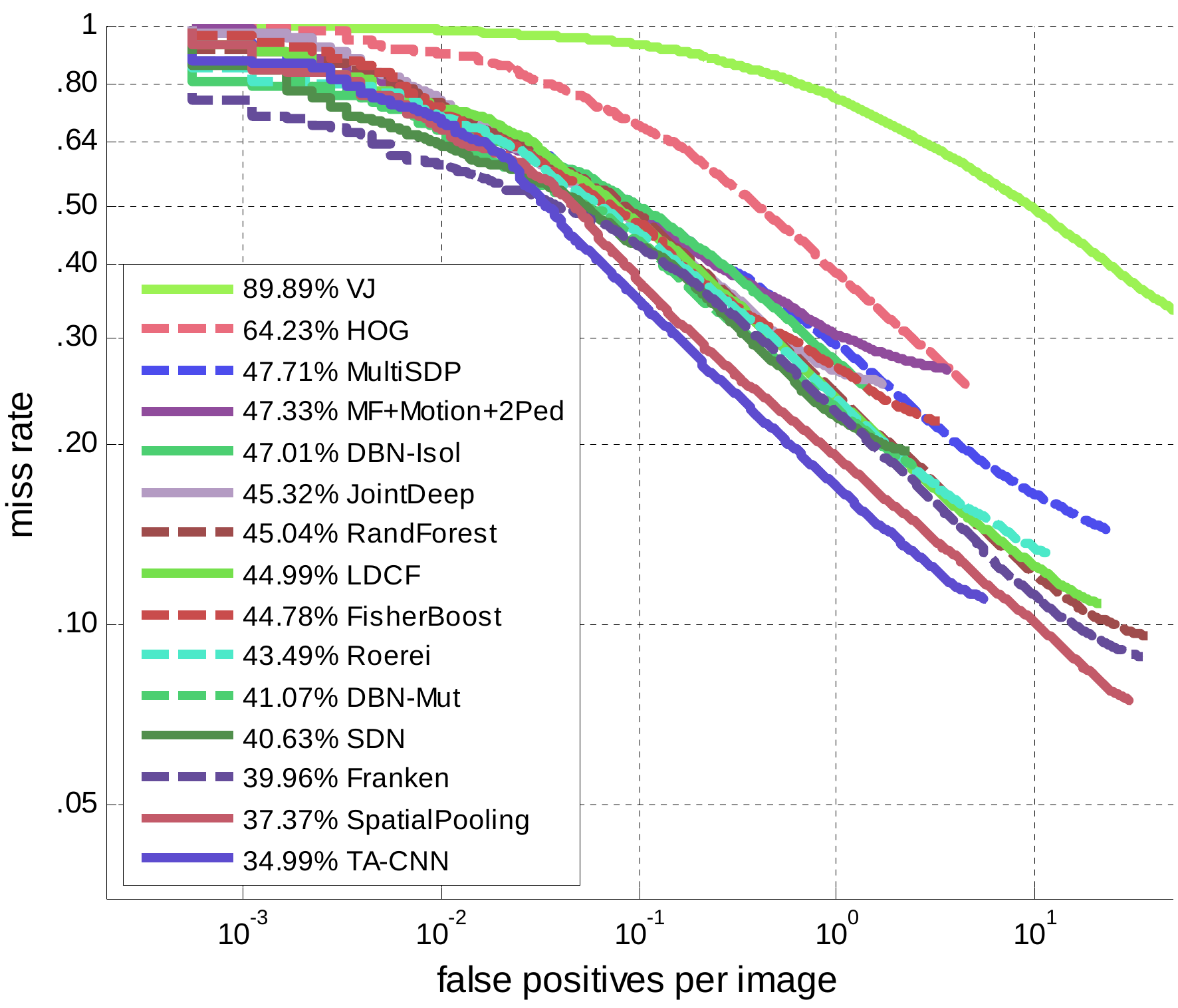}
	\caption{Results on ETH}
	\label{fig:ETH}
\end{figure}

We compare TA-CNN with the existing best-performing methods (see Sec.\ref{sec:caltech})
on ETH \cite{ess2007depth}. TA-CNN is trained on INRIA-Train \cite{DT05}.
This setting aims at evaluating the generalization capacity of the TA-CNN.
As shown in Fig.\ref{fig:ETH}, TA-CNN achieves the lowest miss rate, which outperforms the second best method by $2.5$ percent. It also outperforms the best deep model by $5.5$ percent.

\begin{figure}[t]
	\centering
	\includegraphics[width=0.9\linewidth]{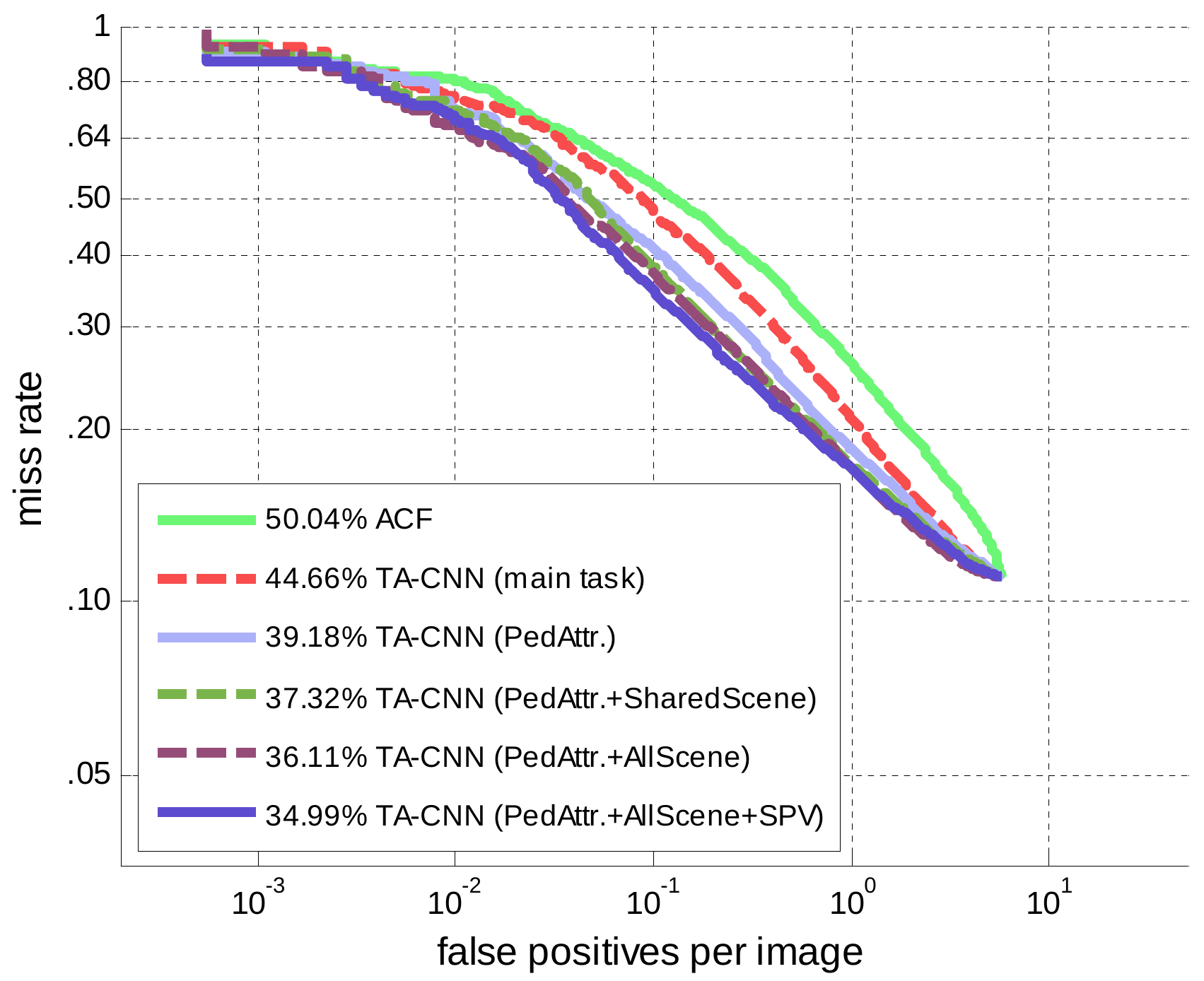}
	\caption{Log-average miss rate reduction procedure on ETH}
	\label{fig:ETH1}
\end{figure}

\textbf{Effectiveness of different Components} The analysis of the effectiveness of different components of TA-CNN is displayed in Fig.\ref{fig:ETH1}, where the log-average miss rates show clear decreasing patterns as follows, while gradually accumulating more components.

$\bullet$ TA-CNN (main task) cascades on ACF and reduces the miss rate by 5.4 percent.

$\bullet$ With pedestrian attributes, TA-CNN (PedAttr.) reduces the result of TA-CNN (main task) by 5.5 percent.

$\bullet$ When bridging the gaps among multiple scene datasets with shared scene attributes, TA-CNN (PedAttr.+SharedScene) further lower the miss rate by 1.8 percent.

$\bullet$ After incorporating unshared attributes, the miss rate is further decreased by another 1.2 percent.

$\bullet$ TA-CNN finally achieves 34.99 percent log-average miss rate with the structure projection vector.

\begin{figure}[t]
	\centering
	\includegraphics[width=0.9\linewidth]{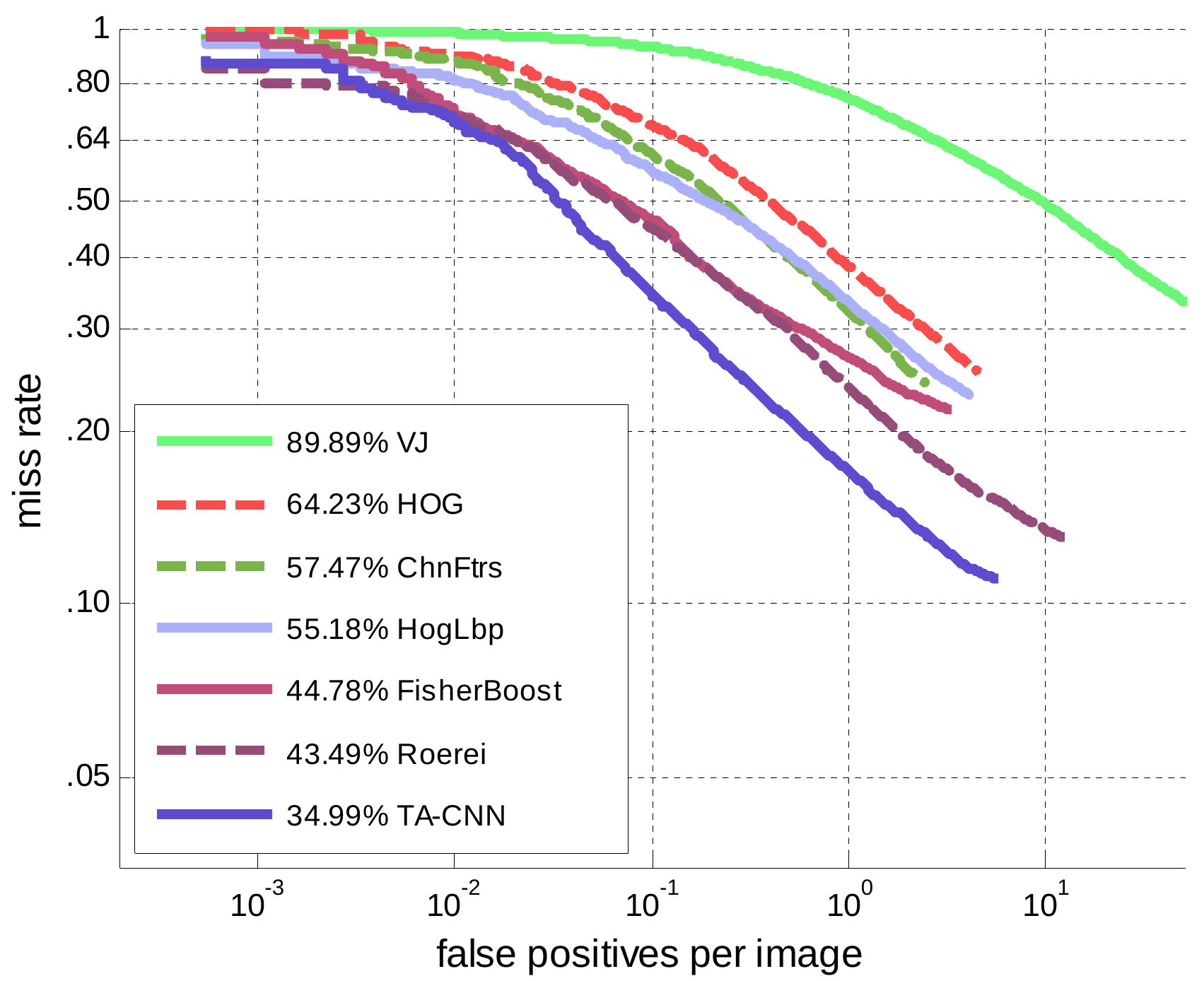}
	\caption{Comparison with hand-crafted feature based models on ETH dataset}
	\label{fig:features}
\end{figure}

\textbf{Comparisons with Hand-crafted Features} Fig.\ref{fig:features} shows that the learned representation of TA-CNN outperforms the conventional handcrafted features in a large margin, including Haar, HOG, HOG+LBP, and channel features. For instance, it reduces the miss rate by 9.8 and 8.5 percent compared to FisherBoost \cite{shen2013training} and Roerei \cite{benenson2013seeking}, respectively. FisherBoost combined HOG and covariance features, and trained the detector in a complex model, while Roerei carefully designed the feature pooling, feature selection, and preprocessing methods based on channel features.

\begin{figure}[t]
	\centering
	\includegraphics[width=0.9\linewidth]{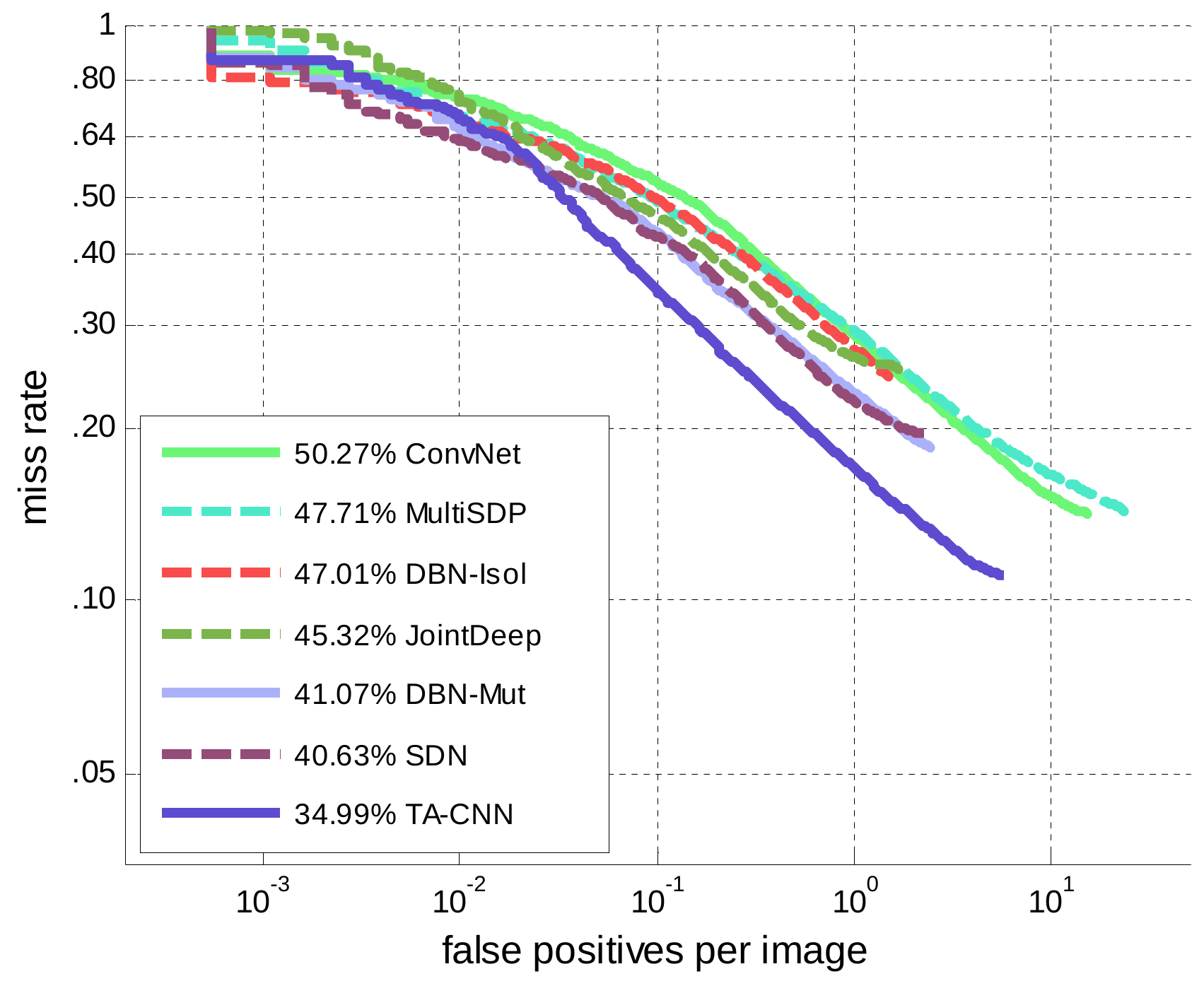}
	\caption{Comparison with other deep models on ETH dataset}
	\label{fig:deepmodels}
\end{figure}

\textbf{Comparisons with Deep Models} Fig.\ref{fig:deepmodels} shows that TA-CNN surpasses other deep models on ETH dataset. For example, TA-CNN outperforms other two best-performing deep models, SDN \cite{luo2013switchable} and DBN-Mul \cite{Ouyang2013MutualDBN}, by 5.5 and 6 percent, respectively. Besides, TA-CNN even reduces the miss rate by 12.7 compared to MultiSDP \cite{zeng2013multi}, which carefully designed multiple classification stages to recognize hard negatives.

\subsection{Visualization of Detection Results}
\begin{figure*}[t]
	\vspace{+15pt}
	\centering
	\includegraphics[width=1.0\textwidth]{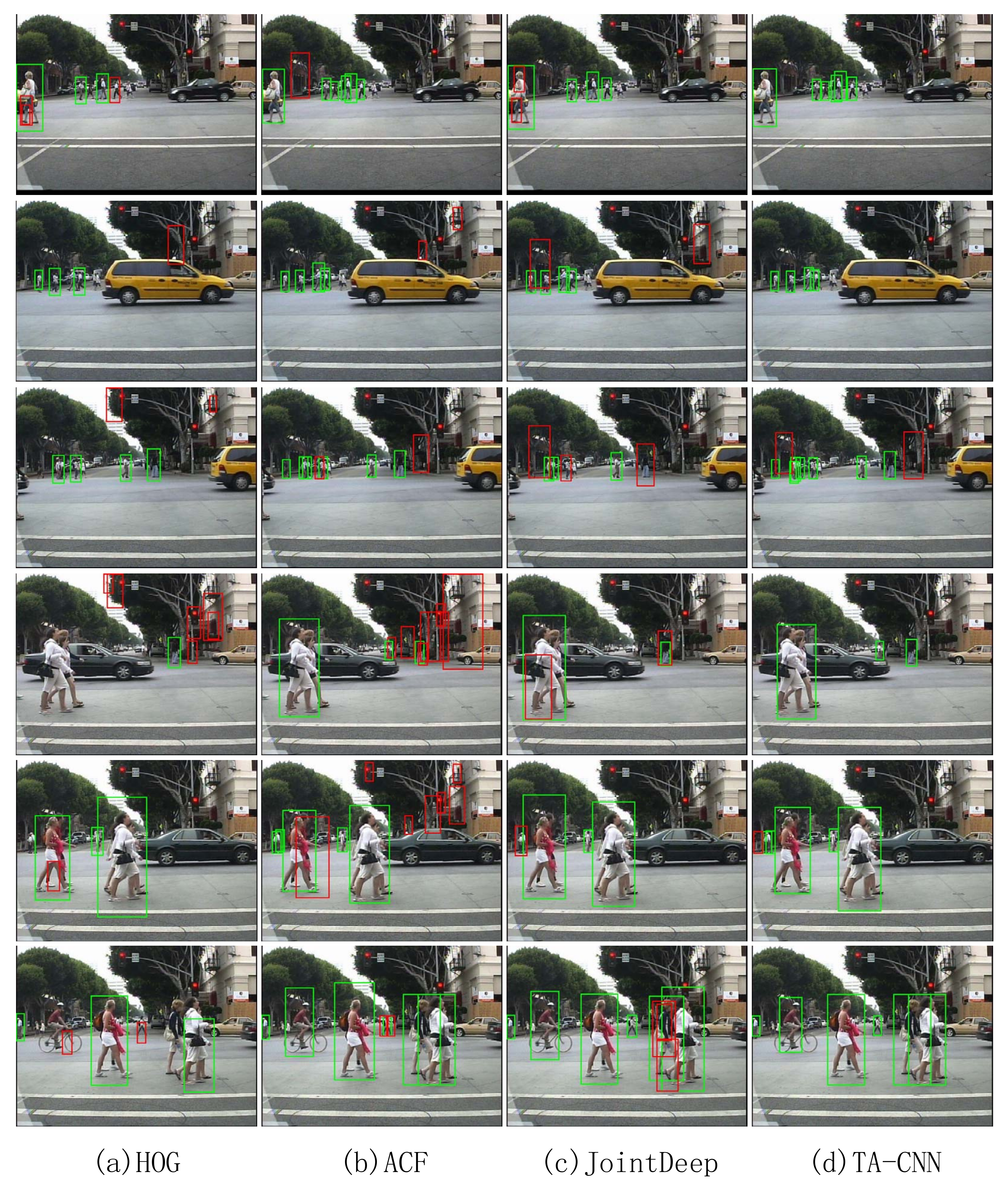}
	\caption{Detection examples of a series of \textbf{continuous crossroad scenes} on \emph{reasonable} subset of Caltech-Test (only consider pedestrians that are larger than $49$ pixels in height and that have $65$ percent visible body parts). Green and red bounding boxes represent true positives and false positives, respectively.}
	\label{fig:Caltech_continuous}\vspace{+15pt}
\end{figure*}

\begin{figure*}[t]
	\vspace{+15pt}
	\centering
	\includegraphics[width=1.0\textwidth]{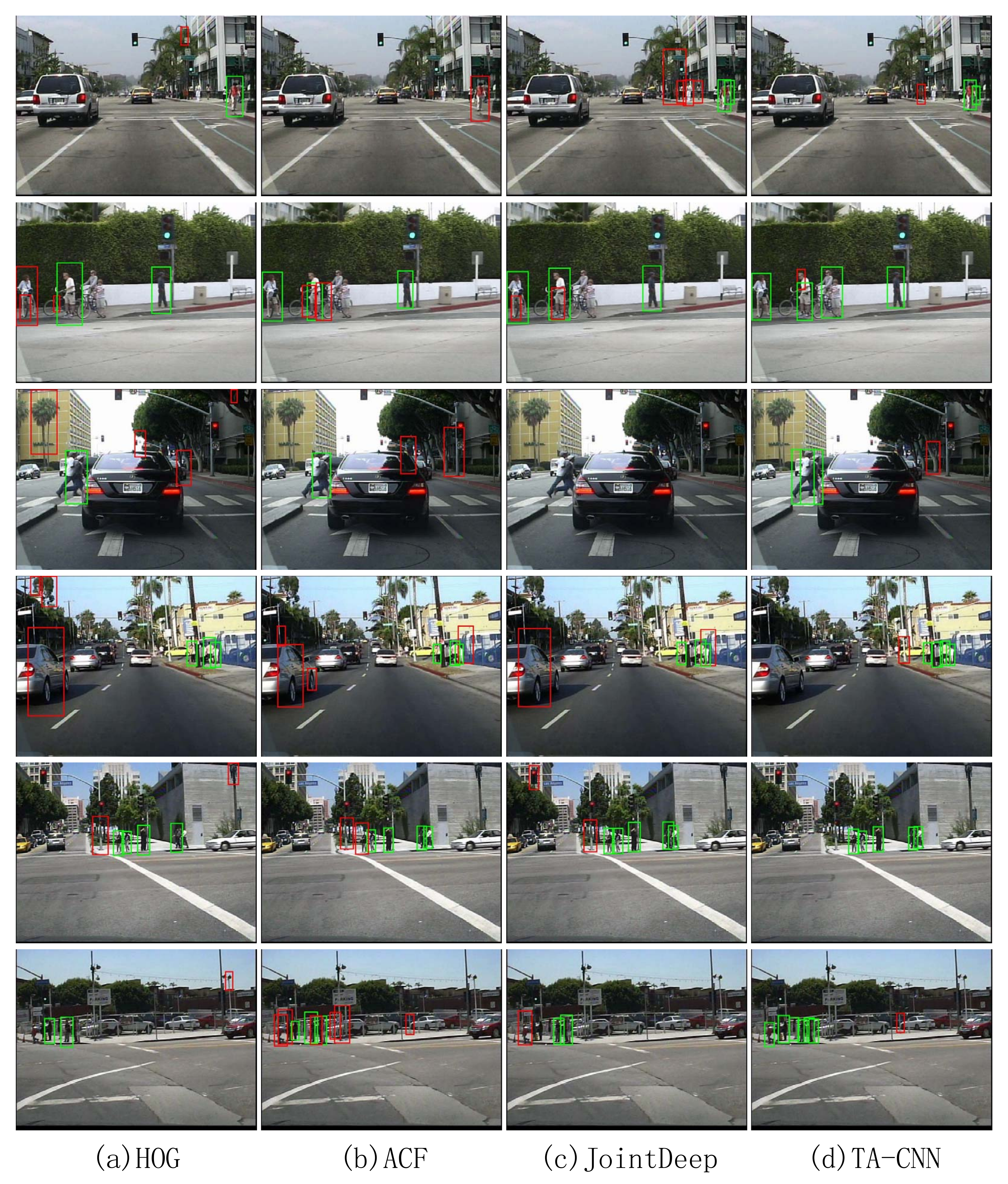}
	\caption{Detection examples on \emph{reasonable} subset of Caltech-Test (only consider pedestrians that are larger than $49$ pixels in height and that have $65$ percent visible body parts). Green and red bounding boxes represent true positives and false positives, respectively.}
	\label{fig:Caltech_discrete}\vspace{+15pt}
\end{figure*}

\begin{figure*}[t]
	\vspace{-15pt}
	\centering
	\includegraphics[width=0.95\textwidth]{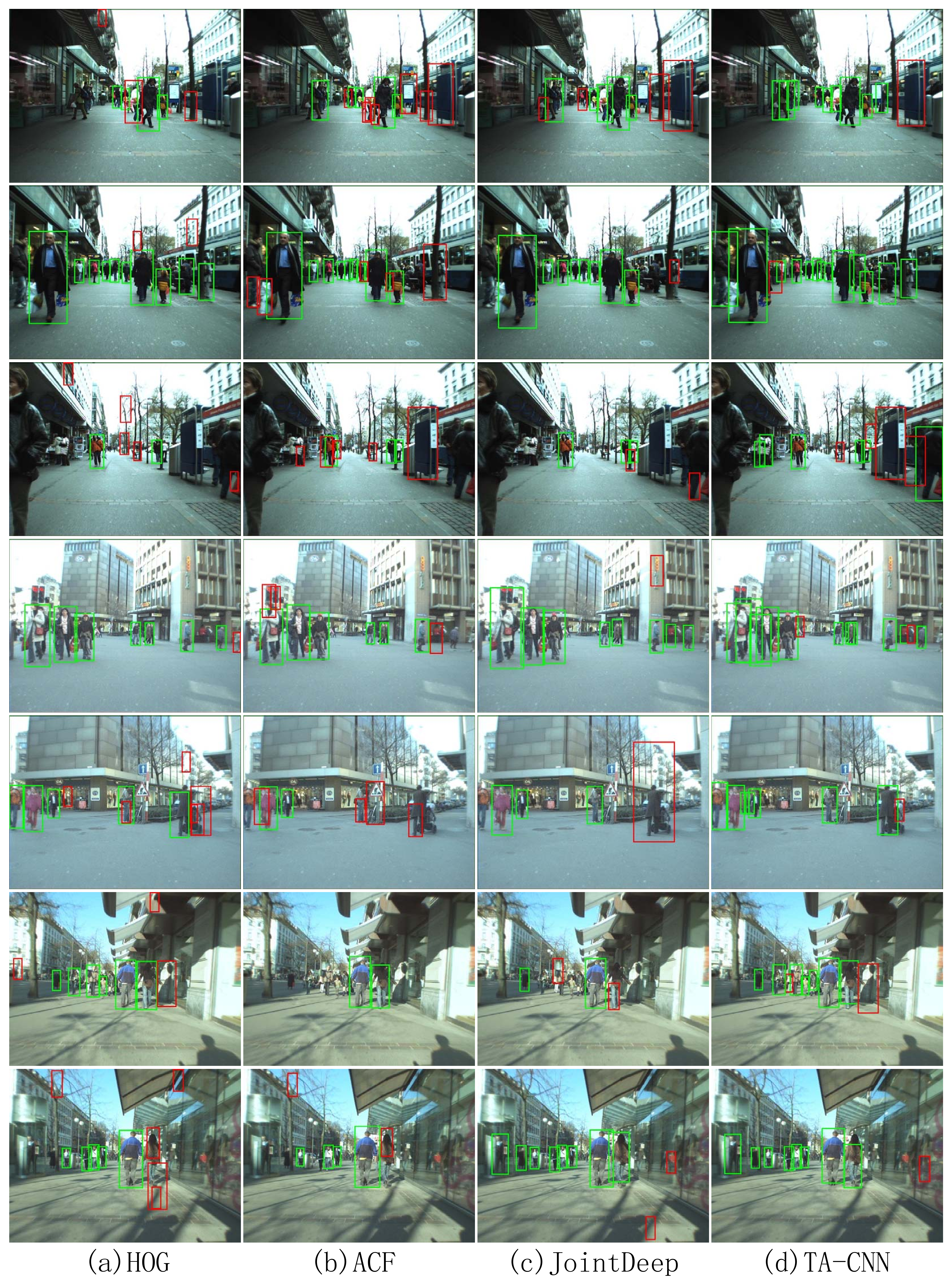}
	\caption{Detection examples on \emph{reasonable} subset of ETH (only consider pedestrians that are larger than $49$ pixels in height and that have $65$ percent visible body parts). Green and red bounding boxes represent true positives and false positives, respectively.}
	\label{fig:ETH_sample}
\end{figure*}

We visualize the results of TA-CNN and compare with HOG \cite{DT05}, ACF \cite{DollarPAMI14pyramids}, and JointDeep \cite{Ouyang2013Joint}. Fig.\ref{fig:Caltech_continuous} and Fig.\ref{fig:Caltech_discrete} show the detection examples on Caltech \emph{reasonable} subset, while Fig.\ref{fig:ETH_sample} shows samples on ETH \emph{reasonable} subset.

\section{Conclusions}

In this paper, we proposed a novel Task-Assistant CNN (TA-CNN) to learn features from multiple tasks (pedestrian and scene attributes) and datasets, showing superiority over hand-crafted features and features learned by other deep models. This is because high-level representation can be learned by employing sematic tasks and multiple data sources. Extensive experiments demonstrate its effectiveness. The proposed model can be further improved by incorporating more attributes. Future work will explore more attribute configurations. The proposed approach also has potential for scene parsing, because it predicts background attributes.

{\small
	\bibliographystyle{ieee}
	\bibliography{egbib}
}

\end{document}